\documentclass{article}
\usepackage{PRIMEarxiv}
\usepackage[utf8]{inputenc}
\usepackage[T1]{fontenc}
\usepackage{booktabs}
\usepackage{graphicx}
\usepackage{subcaption}
\usepackage{amsmath,amssymb,amsthm}
\usepackage{bm}
\usepackage{mathtools}
\usepackage{algorithm}
\usepackage{algpseudocode}
\usepackage{xcolor}
\usepackage{microtype}
\usepackage[numbers,sort&compress]{natbib}
\usepackage{url}
\usepackage[hidelinks]{hyperref}

\theoremstyle{plain}
\newtheorem{theorem}{Theorem}
\newtheorem{lemma}[theorem]{Lemma}
\newtheorem{proposition}[theorem]{Proposition}
\newtheorem{corollary}[theorem]{Corollary}
\theoremstyle{definition}

\newtheorem{assumption}{Assumption}

\newcommand{\E}{\mathbb{E}}
\newcommand{\PP}{\mathbb{P}}
\newcommand{\R}{\mathbb{R}}
\newcommand{\X}{\mathcal{X}}
\newcommand{\A}{\mathcal{A}}
\newcommand{\D}{\mathcal{D}}

\newcommand{\cN}{\mathcal{N}}
\newcommand{\one}{\mathbf{1}}
\DeclareMathOperator*{\argmax}{arg\,max}
\DeclareMathOperator*{\argmin}{arg\,min}
\newcommand{\Var}{\operatorname{Var}}
\newcommand{\OPT}{\mathrm{OPT}}
\newcommand{\Proj}{\Pi}

\newif\ifincludeproofs
\includeproofstrue
\newif\ifincludeneuralntk
\includeneuralntktrue

\hypersetup{
  pdftitle={Action-Centered Causal Bandits for Budget-Constrained Promotion Targeting},
  pdfauthor={Yifan Lin, Kaixu Yang, Di Mo}
}

\title{Action-Centered Causal Bandits for Budget-Constrained Promotion Targeting}

\author{
  Yifan Lin \qquad Kaixu Yang \qquad Di Mo \\
  LinkedIn Corporation \\
  Mountain View, CA, USA \\
  \texttt{\{yiflin,kayang,dmo\}@linkedin.com}
}
\date{}
\begin{document}
\maketitle

\begin{abstract}
Web platforms select among costly interventions for each user to maximize incremental outcomes under a global budget. We formulate this problem as a budget-constrained contextual bandit with heterogeneous treatment effects. Although an action-independent outcome baseline cancels from treatment comparisons, modeling a nonlinear baseline can increase statistical complexity.

We propose DR-CATE-CBwK, an action-centered causal bandit that directly learns treatment contrasts. Its online update remains valid under baseline-model misspecification without inverse-propensity weighting. It combines an offline doubly robust learner to warm-start treatment-effect estimates, an online action-centered estimator for sequential contrast learning, and a primal-dual controller with an affordable-action rule to enforce the effective-cost budget pathwise.

For linear conditional average treatment effects (CATEs) in a stationary i.i.d. setting, we prove a high-probability ($\widetilde{O}(Kd\sqrt{T})$) regret bound, where K is the number of priced actions, d the context dimension, and T the horizon. We benchmark against the fluid linear program (LP) under the coverage-guaranteed action rule. We further show that the action-centered estimating moment remains unbiased under any predictable baseline estimate, whereas a specified separate-arm reward regression can exhibit asymptotic contrast bias under action-dependent logging and shared-baseline misspecification.

On a semi-synthetic job-marketplace benchmark with real logged features and simulated outcomes, our practical Thompson-restricted variant achieves the lowest regret among evaluated budget-feasible methods in the operating regime. It reduces regret by approximately 20\% versus the strongest reward-based baseline, outperforming raw-reward contextual bandits, BwK baselines, and a static offline policy, while never exceeding the effective-cost budget.
\end{abstract}


\section{Introduction}
\label{sec:intro}

Web platforms repeatedly choose, for each user encounter, one of several costly
interventions to maximize an incremental outcome under a global budget: promotional
credits, retention discounts, coupons, or ad subsidies. Candidate actions differ in
cost, their effects differ across users, and total spend is bounded over a horizon.

The standard tool is a contextual bandit
\citep{li2010contextual, agrawal2013thompson} that models the conditional reward
$\E[Y\mid X=x,A=a]=f(x)+\tau_a(x)$: an action-independent baseline $f$ plus the
conditional average treatment effect (CATE) $\tau_a(x)=\E[Y(a)-Y(a_0)\mid X=x]$
against a no-intervention control $a_0$. This decomposition does not change the
objective: for every policy $\pi$,
\begin{equation}
\E_{X,A\sim\pi}[Y(A)]=\E_X[f(X)]+\E_{X,A\sim\pi}[\tau_A(X)],
\label{eq:reward-cate-equivalence}
\end{equation}
and the first term is policy-independent, so a correctly specified reward-based and
a CATE-based bandit with knapsacks (CBwK) share the same constrained optimum. The
difference is statistical: a learner that fits raw conditional rewards also models
the possibly nonlinear nuisance $f$ that cancels from every action comparison. This
can increase estimation variance or, under a misspecified reward class, create
spurious contrasts when separate action models see different context distributions.

A budget makes this distinction consequential because scarce spend must be ranked
by incremental value. If only $20\%$ of traffic can be funded, a \$100 credit with a $+3$
percentage-point uplift on high-baseline power posters but a $+15$ point uplift on
persuadable mid-tier posters should go to the latter, a choice a naive
response-propensity ranking that does not compare treatment against control would
miss. The question is how to learn the treatment contrasts
reliably without first fitting $f$.

We introduce DR-CATE-CBwK, an online algorithm that learns these contrasts under a
hard budget. Building on action-centered bandits
\citep{greenewald2017action,krishnamurthy2018semiparametric}, we couple a
baseline-robust estimator with an offline doubly robust CATE warm start and a
primal-dual controller with an affordable-action restriction. The warm start enters
the confidence radius as an initial estimation error; a valid external radius can
tighten its contribution to regret (Assumption~\ref{ass:offline}). Our contribution
is the joint guarantee for this offline-to-online budgeted learning procedure.

Our contributions are:

(1) DR-CATE-CBwK (Algorithm~\ref{alg:dr-cbwk}), combining offline causal
initialization, online semiparametric contrast learning, and pathwise budget
feasibility. Its farthest-pair action rule supplies active-set coverage without
a permanent exploration floor.

(2) A high-probability $\widetilde O(Kd\sqrt{T})$ regret bound against the fluid LP
for linear CATEs under stationary i.i.d. contexts and the stated noise, coverage,
and confidence conditions (Theorem~\ref{thm:regret}). We also characterize the
baseline robustness of the centered moment and the bias of a specified separate-arm
reward regression (Theorem~\ref{thm:variance}).

(3) A semi-synthetic job-promotion evaluation with real logged features and
simulated outcomes. The practical Thompson-restricted variant has regret $498$
versus $620$ for Raw-Flex-CBwK, the strongest non-ablation baseline, approximately
$20\%$ lower in the operating regime. Comparisons with the farthest-pair and
coverage-mixture variants examine the effect of the action-selection rule on
regret. Additional studies evaluate baseline misspecification, offline-to-online
shift, nonlinear CATEs, and budget feasibility.

\section{Related Work}
\label{sec:related}

A large body of work estimates heterogeneous treatment effects from logged data.
The T- and S-learners fit outcome regressions and difference them; the X-learner
\citep{kunzel2019metalearners} regresses imputed treatment effects and can exploit
treatment imbalance; the R-learner \citep{nie2021quasi} orthogonalizes against a
baseline nuisance via Robinson's transformation; and the DR-learner
\citep{kennedy2023towards, dudik2014doubly} uses the augmented inverse-propensity
weighted pseudo-outcome and inherits doubly-robust consistency. Neyman
orthogonality reduces first-order sensitivity to nuisance estimation
\citep{foster2023orthogonal}; Assumption~\ref{ass:nuisance} specifies the
asymptotic nuisance product-rate condition used for our offline learner. Policy
learning from doubly-robust scores \citep{athey2021policy} is the closest
precedent for our offline warm start. We use action-specific DR pseudo-outcomes
to initialize the CATE estimates from non-uniformly logged data. The online
action-centered update then learns the same treatment contrasts from sequential
interaction.

LinUCB \citep{li2010contextual,chu2011contextual}, linear Thompson sampling
\citep{agrawal2013thompson}, and regression-oracle contextual bandits
\citep{simchilevi2022bypassing} target conditional reward. Under our
decomposition $\E[Y\mid X,A=a]=f(X)+\tau_a(X)$, that population objective has
the same optimizer as incremental value (Eq.~\eqref{eq:reward-cate-equivalence});
the difference is that a standard linear-reward realization must represent the
nuisance $f$, whereas the semiparametric model need only realize the contrasts.
\citet{greenewald2017action} introduced action-centered contextual bandits,
separating a complex baseline reward from a simple treatment effect and using
action-centering to learn the latter without online inverse-propensity weighting.
\citet{krishnamurthy2018semiparametric} developed BOSE for action-independent
confounding and obtained the $\widetilde O(d\sqrt T)$ rate without modeling $f$;
its confidence analysis is the basis of our regret argument.

Online resource allocation under budgets is formalized as bandits with knapsacks
\citep{badanidiyuru2018bandits}, with contextual algorithms based on policy
optimization \citep{agrawal2016efficient} and adaptive linear programming
\citep{wu2015algorithms}. A primal-dual approach uses
online convex optimization over a dual variable atop a
per-step budget-adjusted argmax, with adaptive pacing \citep{balseiro2019learning};
recent work refines the contextual frontier \citep{slivkins2023contextual,
chen2024contextual, guo2025stochastic}. These results allow generic reward
signals but do not combine a BOSE-style arbitrary action-independent nuisance
with an offline causal warm start and a hard spending constraint.

A separate offline line targets cost-aware uplift directly: \citet{zhao2019uplift}
extend meta-learners to optimize net incremental value across treatments with
different costs. This motivates comparison with a static allocation policy.
Our Offline-DR baseline fixes the offline treatment-effect estimates and a
budget-tuned dual (Section~\ref{sec:exp-methods}), while DR-CATE-CBwK updates
both online.

Closest in application are recent online causal and budgeted promotion systems.
\citet{sawant2018contextual} combine uplift targeting with contextual Thompson
sampling but without a budget or regret analysis, and DISCO \citep{zhang2024disco}
allocates discounts under a global operational constraint with a live A/B test but
targets a reward model rather than the CATE and gives no regret guarantee. Relative
to these systems we add an offline causal warm start, remove an arbitrary shared
baseline through action-centering, handle multiple priced actions, and prove a
hard-budget regret bound. This coupling provides a semiparametric offline-to-online
guarantee for budgeted promotion targeting. On the design side,
\citet{kim2025experimental} study experimental-design semiparametric bandits,
relevant to our active-set coverage rule.

Closest to our setting, \citet{kim2021doubly} develop doubly-robust Thompson
sampling for linear contextual bandits, applying the correction to the raw reward
rather than an incremental contrast, and without a budget.
\citet{dimakopoulou2019balanced} incorporate causal balancing methods into
reward estimation for linear contextual bandits, and \citet{bibaut2021post, hadad2021confidence,
zhan2021offpolicy} study valid inference after adaptive data collection, motivating
our explicit propensity bookkeeping. Our online update builds on action-centered
and semiparametric bandits
\citep{greenewald2017action,krishnamurthy2018semiparametric}. The distinction is
the joint guarantee with an offline doubly robust warm start and a primal-dual
hard-budget controller.

\section{Problem Setting}
\label{sec:setup}

\subsection{CATE Under Potential Outcomes}

For decision epoch $t = 1, \ldots, T$: $X_t \in \X \subseteq \R^d$ is the observed
context; $A_t \in \A := \{a_0, a_1, \ldots, a_K\}$ is the action, with $a_0$ the
no-promotion control; $Y_t \in \R$ is the observed outcome;
$c : \X \times \A \to \R_{\geq 0}$ is a known per-decision cost with
$c(\cdot, a_0) \equiv 0$; and $B \in \R_{>0}$ is the budget over the horizon, with
per-period rate $\rho := B / T$.

\begin{assumption}[i.i.d.\ contexts]\label{ass:iid}
Contexts $X_t$ are drawn i.i.d.\ from a fixed distribution $P_X$.
\end{assumption}

\begin{assumption}[Structured outcomes and bounded costs]\label{ass:bounded}
The outcome decomposes as
$Y_t = f(X_t) + \tau_{A_t}(X_t) + \xi_t$, where $f(x)=\E[Y\mid X=x,A=a_0]$ is an
action-independent baseline with $|f(x)|\le F$, the treatment contrasts
$\tau_a$ are the CATEs defined below, and the bounded noise $\xi_t$ is mean-zero and
independent of the action given the context and history
(Assumption~\ref{ass:subgauss}). Costs
satisfy $c(x, a) \in [0, c_{\max}]$ with
$c(x, a) \geq c_{\min} > 0$ for $a \neq a_0$.
\end{assumption}

\begin{assumption}[Bounded contexts and parameters]\label{ass:norms}
$\|X_t\|_2 \le L$ almost surely, and the stacked treatment parameter satisfies
$\|\theta^\star\|_2\le S$.
\end{assumption}

We adopt the Neyman-Rubin potential-outcomes framework, in which unit $t$ has
potential outcomes $\{Y_t(a): a \in \A\}$ and only $Y_t = Y_t(A_t)$ is observed.
The conditional average treatment effect of action $a$ against the control is
\begin{equation*}
  \tau_a(x) := \E\!\left[\, Y(a) - Y(a_0) \,\middle|\, X = x \,\right].
\end{equation*}
Under Assumptions~\ref{ass:norms} and~\ref{ass:realizability},
$\tau_a(x)=x^\top\theta_a^\star \in [-R_{\max}, R_{\max}]$ with
$R_{\max} := L S$. Identification of $\tau_a$ and unbiased off-policy
evaluation rely on consistency and no interference (SUTVA), overlap, and
unconfoundedness. For the offline log, overlap means
$e(a\mid x):=\PP(A=a\mid X=x)\ge\epsilon_{\mathrm{off}}>0$ for every action
on the logged context support.

The two phases invoke unconfoundedness differently. Offline, unconfoundedness given
the recorded context, $\{Y_t(a)\}_{a\in\A}\perp A_t\mid X_t$, is a maintained
assumption, as in observational policy learning \citep{athey2021policy}.
Online it holds by design: with history
$\mathcal{H}_{t-1}:=\{(X_s,A_s,Y_s)\}_{s<t}$, $A_t$ is drawn from the algorithm's
known law $\pi_t$, so $\{Y_t(a)\}_{a\in\A}\perp A_t\mid X_t,\mathcal{H}_{t-1}$ and the
propensity $P_t=\pi_t(A_t\mid X_t,\mathcal H_{t-1})$ is known by construction.

\begin{assumption}[Product-of-rates on nuisances]\label{ass:nuisance}
Let $\mu(x,a):=\E[Y\mid X=x,A=a]$ and $e(a\mid x):=\PP(A=a\mid X=x)$, and let
$\hat\mu,\hat e$ be cross-fitted estimators over an offline sample of size
$n_{\mathrm{off}}$. Then
$\|\hat\mu-\mu\|_{L^2(P_X)}\,\|\hat e-e\|_{L^2(P_X)}=o(n_{\mathrm{off}}^{-1/2})$.
\end{assumption}

The regret analysis assumes a linearly realizable CATE; nonlinear effects are
evaluated empirically in Section~\ref{sec:exp-nonlinear}.

\begin{assumption}[Linear realizability of the CATE]\label{ass:realizability}
There exists $\theta_a^\star\in\R^d$ with $\tau_a(x)=x^\top\theta_a^\star$ for all
$a\in\A$ and all $x$ in the support of $P_X$.
\end{assumption}

\subsection{Contextual Bandit With Budget and Regret}

A policy $\pi : \X \to \Delta(\A)$ induces incremental value and cost
\begin{equation}
  V(\pi) := \E_X \E_{A \sim \pi(\cdot \mid X)}[\, \tau_A(X) \,], \quad
  C(\pi) := \E_X \E_{A \sim \pi(\cdot \mid X)}[\, c(X, A) \,].
  \label{eq:vc}
\end{equation}
The incremental objective only reparameterizes the reward objective: for every
feasible $\pi$,
\begin{equation}
\E_{X,A\sim\pi}[Y(A)]=\E_X[f(X)]+V(\pi),
\label{eq:policy-equivalence}
\end{equation}
so the constant $\E[f(X)]$ changes neither feasibility nor the maximizer, which we
learn without requiring $f$ to lie in the online linear class.
We target the rate-based program $\max_\pi V(\pi)$ subject to $C(\pi)\le\rho$,
with Lagrangian
\begin{equation}
  L(\pi, \lambda) = \E_X \E_{A \sim \pi(\cdot \mid X)}\big[\, \tau_A(X) - \lambda\, c(X, A) \,\big] + \lambda \rho.
  \label{eq:lagrangian}
\end{equation}
At fixed $\lambda \geq 0$ the inner maximizer is pointwise,
\begin{equation}
  \pi_\lambda^\star(x) \in \argmax_{a \in \A} \big(\, \tau_a(x) - \lambda\, c(x, a) \,\big),
  \label{eq:pi-lambda}
\end{equation}
which motivates the online budget-adjusted score. The fluid LP relaxes the
pathwise spending constraint to an expected-cost constraint and supplies the
benchmark for cumulative pseudo-regret:
\begin{equation*}
  R(T) := T \cdot \OPT_{\text{LP}}(\rho) - \sum_{t=1}^T \tau_{A_t}(X_t),
  \quad \OPT_{\text{LP}}(\rho) := \max_{\pi:\,C(\pi)\le\rho} V(\pi).
\end{equation*}

\section{Algorithm}
\label{sec:algorithm}

DR-CATE-CBwK has an offline stage that produces a warm-start CATE estimate from
a cross-fitted doubly robust learner, and an online stage that, each round, draws a
randomized budget-aware action, forms an action-centered feature, and updates a
single regularized least-squares estimate of the linear CATE together with a
primal-dual budget variable. Two sub-procedures are invoked by the main loop: the
offline warm start (Algorithm~\ref{alg:offline-dr}) and the randomized
budget-aware action (Algorithm~\ref{alg:sample-mix}).

\subsection{Offline Warm Start}

We use a doubly robust learner offline to initialize the online estimator. On an
offline dataset split into $K_{\mathrm{fold}}$ folds (with $k(i)$ the fold of
observation $i$) and nuisances $\hat\mu^{(-k)},\hat e^{(-k)}$ trained outside fold
$k$, the cross-fitted DR pseudo-outcome for action $a$ against the control is
\begin{align}
  \psi_i(a) =\; & \hat\mu^{(-k(i))}(X_i, a) - \hat\mu^{(-k(i))}(X_i, a_0) \nonumber\\
  & + \tfrac{\one\{A_i = a\}}{\hat e^{(-k(i))}(a \mid X_i)}\big(Y_i - \hat\mu^{(-k(i))}(X_i, a)\big) \nonumber\\
  & - \tfrac{\one\{A_i = a_0\}}{\hat e^{(-k(i))}(a_0 \mid X_i)}\big(Y_i - \hat\mu^{(-k(i))}(X_i, a_0)\big).
  \label{eq:dr-pseudo}
\end{align}
When the logging propensities $P_i$ are recorded, as in our setting, $\hat e$ is
set to the known logging law rather than estimated; otherwise it is cross-fitted,
subject to the product-rate condition in Assumption~\ref{ass:nuisance}.
Regressing $\psi_i(a)$ on $X_i$ and stacking the arms gives a warm-start mean
$\hat\theta^{(0)}$ that seeds the online estimator's prior mean; since
$V_0=\alpha I$, the offline sample initializes the mean while the online design
starts isotropically. The outcome model also supplies the clipped baseline
$\hat f(x):=\Proj_{[-F,F]}\hat\mu(x,a_0)$ (used online to residualize), with
$|f-\hat f|\le2F$ whenever $|f|\le F$. By Neyman orthogonality the nuisance product
enters $\hat\theta^{(0)}$'s procedure-specific bounds only at second order
(Assumption~\ref{ass:nuisance}); the online theorem instead accepts any valid prior
radius, including the deterministic $2S$.

\subsection{Online Action-Centered Update}

We write the outcome as
\begin{equation}
Y_t = f(X_t) + \tau_{A_t}(X_t) + \xi_t,\qquad \tau_a(x)=x^\top\theta_a^\star,
\label{eq:outcome-model}
\end{equation}
where $f(x)=\mu(x,a_0)$ is an arbitrary, possibly nonlinear, action-independent
baseline and $\xi_t$ is mean-zero and bounded given $(X_t,\mathcal H_{t-1})$ and
independent of the action $A_t$
(Assumption~\ref{ass:subgauss}). Stack the per-arm parameters into
$\theta^\star=(\theta_{a_1}^\star,\dots,\theta_{a_K}^\star)\in\R^{Kd}$;
the control has $\theta_{a_0}^\star=0$ and is omitted from the stack. Define the
block action feature and its mean under the known
online action law $\pi_t$,
\begin{equation}
z_t(a)=e_a\otimes X_t\in\R^{Kd},\qquad
\bar z_t=\sum_{a\in\A_t^{\mathrm{aff}}}\pi_t(a)\,z_t(a),
\label{eq:block-feature}
\end{equation}
where $e_{a_j}$ is the $j$th standard basis vector of $\R^K$ and $e_{a_0}=0$,
so that $z_t(a)^\top\theta^\star=\tau_a(X_t)$. Here
$\A_t^{\mathrm{aff}}:=\{a\in\A:c(X_t,a)\le B-S_{t-1}\}$ is the affordable set of
actions the residual budget $B-S_{t-1}$ can fund, with $S_{t-1}=\sum_{s<t}c(X_s,A_s)$
the cumulative spend; it always contains the cost-free control $a_0$. We abbreviate
$\pi_t(a):=\pi_t(a\mid X_t,\mathcal H_{t-1})$. The online update uses the
action-centered pair $(u_t,g_t)$,
\begin{equation}
u_t=z_t(A_t)-\bar z_t,\qquad g_t=Y_t-\hat f(X_t).
\label{eq:online-dr}
\end{equation}
Because $\bar z_t$ is the exact conditional mean of $z_t(A_t)$ under $\pi_t$, we have
$\E[u_t\mid X_t,\mathcal H_{t-1}]=0$, so the arbitrary baseline, and any error
$f-\hat f$ in its estimate, cancels in expectation:
\begin{equation}
\E[u_t\,g_t\mid X_t,\mathcal H_{t-1}]=\E[u_t u_t^\top\mid X_t,\mathcal H_{t-1}]\,\theta^\star=:M_t\,\theta^\star.
\label{eq:online-unbiased-main}
\end{equation}
This action-centering identity
\citep{greenewald2017action,krishnamurthy2018semiparametric} targets the contrasts with no
inverse-propensity weight; the score $u_tg_t$ is unbiased
for any predictable $\hat f$, including $\hat f\equiv0$, so the baseline affects only
the variance (Theorem~\ref{thm:variance}(c)). Clipping $\hat f$ to $[-F,F]$ gives
$\|u_tg_t\|_2\le2L(2F+LS+\Xi)$. We maintain the
regularized least-squares estimate
\begin{equation}
V_t=V_{t-1}+u_t u_t^\top,\qquad
\tilde\theta_t=V_t^{-1}\Big(\alpha\hat\theta^{(0)}+\textstyle\sum_{s\le t}u_s g_s\Big),
\label{eq:ls-update}
\end{equation}
with $V_0=\alpha I$, warm-started at $\hat\theta^{(0)}$. Before scoring, we set
$\hat\theta_t$ to the projection of $\tilde\theta_t$
onto $\{\theta:\|\theta\|_2\le S\}$ in the $V_t$ norm, which cannot enlarge the
$V_t$-error since $\theta^\star$ lies in the set. Because $u_t$ depends on the
sampled action, the confidence set is the semiparametric
self-normalized bound of Section~\ref{sec:theory}.

\begin{algorithm}[t]
  \caption{DR-CATE-CBwK: causal contextual bandit with an action-centered update
  and primal-dual budget control.}
  \label{alg:dr-cbwk}
  \begin{algorithmic}[1]
    \Require Budget $B$, horizon $T$, logged dataset $\D$, dual step $\eta_0$,
    confidence radius $\beta_t$ (set in Theorem~\ref{thm:regret}).
    \State $(\hat\theta^{(0)}, \hat f, \lambda_1) \gets \textsc{OfflineDR}(\D)$
    \Comment{Algorithm~\ref{alg:offline-dr}}
    \State $V_0 \gets \alpha I$;\ \ $b_0\gets \alpha\hat\theta^{(0)}$;\ \ $\hat\theta_0 \gets \hat\theta^{(0)}$;\ \ $S_0 \gets 0$
    \For{$t = 1, \ldots, T$}
      \State Observe $X_t$.
      \State $\A_t^{\mathrm{aff}} \gets \{a \in \A : c(X_t, a) \le B - S_{t-1}\}$
      \Comment{always contains $a_0$}
      \State $(\pi_t, A_t) \gets \textsc{RandomizedAction}(\hat\theta_{t-1}, V_{t-1}, X_t, \lambda_t, \A_t^{\mathrm{aff}}; \beta_t)$
      \State Observe $Y_t$; incur $c(X_t, A_t)$; $S_t \gets S_{t-1} + c(X_t, A_t)$.
      \State $\bar z_t \gets \sum_{a\in\A_t^{\mathrm{aff}}}\pi_t(a)\,z_t(a)$;\ \
      $u_t \gets z_t(A_t)-\bar z_t$;\ \ $g_t \gets Y_t-\hat f(X_t)$.
      \State $V_t \gets V_{t-1}+u_tu_t^\top$;\ \ $b_t\gets b_{t-1}+u_tg_t$;\ \
      $\hat\theta_t \gets \argmin_{\|\theta\|_2\le S}\|\theta-V_t^{-1}b_t\|_{V_t}$.
      \State $\lambda_{t+1} \gets \big[\lambda_t + \eta_t (c(X_t, A_t) - \rho)\big]_+$
    \EndFor
  \end{algorithmic}
\end{algorithm}

\begin{algorithm}[t]
  \caption{\textsc{OfflineDR}: cross-fitted doubly robust warm start.}
  \label{alg:offline-dr}
  \begin{algorithmic}[1]
    \Require Logged $\D = \{(X_i, A_i, P_i, Y_i)\}$, folds $K_{\mathrm{fold}}$, ridge $\alpha$.
    \For{$k = 1, \ldots, K_{\mathrm{fold}}$} fit $\hat\mu^{(-k)}$ (and $\hat e^{(-k)}$, unless propensities are logged) outside fold $k$. \EndFor
    \State Compute $\psi_i(a)$ for each $i,a$ via Eq.~\eqref{eq:dr-pseudo}.
    \State Regress $\psi_i(a)$ on $X_i$ to get the prior mean $\hat\theta^{(0)}$; set
    baseline $\hat f\gets\Proj_{[-F,F]}\hat\mu(\cdot,a_0)$; project
    $\hat\theta^{(0)}$ onto the radius-$S$ ball; init
    $\lambda_1\in[0,R_{\max}/c_{\min}]$ by the offline Lagrangian.
    \State \Return $(\hat\theta^{(0)}, \hat f, \lambda_1)$.
  \end{algorithmic}
\end{algorithm}

\begin{algorithm}[t]
  \caption{\textsc{RandomizedAction}: active-set semiparametric selection with a
  budget offset (Assumption~\ref{ass:info}).}
  \label{alg:sample-mix}
  \begin{algorithmic}[1]
    \Require Estimate $\hat\theta$, precision $V$, context $X_t$, dual $\lambda_t$,
    affordable set $\A_t^{\mathrm{aff}}$; confidence radius $\beta_t$ (Theorem~\ref{thm:regret}).
    \State $\hat s_t(a) \gets z_t(a)^\top\hat\theta - \lambda_t c(X_t,a)$
    for $a\in\A_t^{\mathrm{aff}}$.
    \State Form the active set by pairwise elimination,
    \begin{equation}
    \mathcal B_t=\big\{a\in\A_t^{\mathrm{aff}} :\ \hat s_t(a')-\hat s_t(a)\le 2\beta_t\|z_t(a')-z_t(a)\|_{V^{-1}}\ \ \forall a'\in\A_t^{\mathrm{aff}}\big\}.
    \label{eq:active-set}
    \end{equation}
    \State If $|\mathcal B_t|=1$, put unit mass on its action. Otherwise choose
    $(a_t^+,a_t^-)\in\argmax_{a,a'\in\mathcal B_t,\,a\ne a'}
    \|z_t(a)-z_t(a')\|_{V^{-1}}$ and put probability $1/2$ on each endpoint.
    \State Sample $A_t \sim \pi_t$; record $\pi_t$.
    \State \Return $(\pi_t, A_t)$.
  \end{algorithmic}
\end{algorithm}

\subsection{The Action-Centered Estimator and Information Accumulation}

The estimator in Eq.~\eqref{eq:ls-update} accumulates information only through the
centered features $u_t$. At a fixed context the single-round information
$M_t=\mathrm{Cov}(e_{A_t})\otimes X_tX_t^\top$ has rank at most $K$, hence is singular
in $\R^{Kd}$ for $d>1$, so information must accumulate across contexts.
Algorithm~\ref{alg:sample-mix} is therefore an active-set semiparametric rule
\citep{krishnamurthy2018semiparametric}: it eliminates every action beaten beyond its
joint confidence width, leaving the plausible set $\mathcal B_t$ of
Eq.~\eqref{eq:active-set}, and samples within $\mathcal B_t$ to meet the
optimal-design condition Eq.~\eqref{eq:coverage}. Imposing that condition only on the
active set, not as a uniform floor, lets $\mathcal B_t$ collapse to a singleton once a
budget-adjusted gap resolves, stopping exploration with no linear-regret term. The
farthest-pair rule examines $O(|\mathcal B_t|^2)$ pairs and attains
$\kappa_0=1/4$: if $\Delta_t=z_t(a_t^+)-z_t(a_t^-)$, each centered endpoint
is $\pm\Delta_t/2$, so the expected squared centered norm is
$\|\Delta_t\|_{V_{t-1}^{-1}}^2/4$, one quarter of the maximum pairwise distance
squared. A mixture assigning weight $\gamma>0$ to this rule has covariance at
least $\gamma$ times the farthest-pair covariance and hence coverage
$\kappa_0=\gamma/4$. The practical Thompson-restricted alternative need not
satisfy this condition. All three rules use the same estimator and affordable set;
only rules satisfying coverage with the prescribed confidence parameters carry
the regret guarantee.

\subsection{Primal-Dual Budget Control}

To adapt the budget price as online estimates improve, we use the nonnegative
virtual-queue update
\begin{equation}
  \lambda_{t+1} := \big[\lambda_t + \eta_t (c(X_t, A_t) - \rho)\big]_+,
  \label{eq:dual-update}
\end{equation}
with step $\eta_t = \eta_0/\sqrt{T}$. Projection only at zero gives a
prefix-spend inequality by telescoping the queue. The active-set rule also bounds
$\lambda_t$: once the price exceeds every plausible treatment gain, only the free
control remains active and the queue decreases. Thus the update achieves
$\widetilde O(\sqrt T)$ dual regret under the theorem's conditions while responding
to deviations from the target spending rate $\rho$. The affordable-set
restriction $\A_t^{\mathrm{aff}}$ confines each round's
sampling to actions the residual budget can fund and always retains the cost-free
control arm, so spending falls back to $a_0$ once no priced action is affordable;
by induction, $S_t\le S_{t-1}+(B-S_{t-1})=B$ for every round. This pathwise
guarantee does not depend on the confidence calibration or coverage condition.
An optional uniform exploration floor plays
$\pi_t^{\mathrm{floor}}=(1-\epsilon_{\mathrm{floor}})\pi_t
+\epsilon_{\mathrm{floor}}\mathrm{Unif}(\A_t^{\mathrm{aff}})$.
Theorem~\ref{thm:regret} analyzes the default setting
$\epsilon_{\mathrm{floor}}=0$. A positive floor preserves pathwise affordability
but changes the queue and regret analysis.

\section{Theoretical Analysis}
\label{sec:theory}

For the coverage-guaranteed instantiation of Algorithm~\ref{alg:dr-cbwk}, regret
decomposes into online bandit regret, dual regret, and a warm-start term. In
addition to Assumptions~\ref{ass:iid}, \ref{ass:bounded}, \ref{ass:norms}, and
\ref{ass:realizability}, the analysis uses bounded action-independent noise,
active-set coverage, and an offline confidence condition. Context diversity is
used separately in the identification discussion of Theorem~\ref{thm:variance}.
Throughout, $K$ is the number of priced arms, so the stacked
parameter $\theta^\star\in\R^{Kd}$ has effective dimension $Kd$.

\begin{assumption}[Bounded action-independent noise]\label{ass:subgauss}
Conditional on $(X_t,\mathcal H_{t-1})$, the noise $\xi_t$ is mean-zero, satisfies
$|\xi_t|\le\Xi$ almost surely, and is independent of the sampled action $A_t$. In
particular it is conditionally $\Xi$-sub-Gaussian.
\end{assumption}

Unbiasedness of the action-centered score follows from the construction and the
known online law (Eq.~\eqref{eq:online-unbiased-main}) for any predictable $\hat f$.
Since $u_t$ depends on the sampled action, we use the semiparametric
self-normalized inequality of \citet{krishnamurthy2018semiparametric}. It requires
the scalar residual to be
conditionally independent of $u_t$ given $(X_t,\mathcal H_{t-1})$, and
Assumption~\ref{ass:subgauss} secures exactly this: the residual
$\zeta_t=f(X_t)-\hat f(X_t)+\bar z_t^\top\theta^\star+\xi_t$ has a predictable
non-noise part and an action-independent noise $\xi_t$, so
$\zeta_t$ is conditionally independent of $u_t$ given $(X_t,\mathcal H_{t-1})$.
No inverse-propensity weight enters, and clipping
the baseline to its known range supplies the bounded residual it needs. The regret
proof adopts this action-independent-noise condition of BOSE; the centered moment
remains unbiased under action-dependent conditionally mean-zero noise, such as the
action-dependent variance of a Bernoulli conversion.

\begin{assumption}[Context diversity]\label{ass:diversity}
There is a constant $\sigma_X^2>0$ with
$\E[X_tX_t^\top\mid\mathcal H_{t-1}]\succeq\sigma_X^2 I$ for all $t$.
\end{assumption}

\begin{assumption}[Active-set semiparametric selection]\label{ass:info}
With the confidence-based active set $\mathcal B_t$ of Eq.~\eqref{eq:active-set},
the rule of Algorithm~\ref{alg:sample-mix} draws a predictable distribution $\pi_t$
supported on $\mathcal B_t$ that satisfies the semiparametric optimal-design
condition: for a constant $\kappa_0>0$ and every pair $a,a'\in\mathcal B_t$,
\begin{equation}
\|z_t(a)-z_t(a')\|_{V_{t-1}^{-1}}^2
\le \kappa_0^{-1}\,\E_{A\sim\pi_t}\big[\|z_t(A)-\bar z_t\|_{V_{t-1}^{-1}}^2\,\big|\,X_t,\mathcal H_{t-1}\big].
\label{eq:coverage}
\end{equation}
\end{assumption}

Since $\mathrm{rank}(M_t)\le K$, the single-round information matrix is singular
for $d>1$. Coverage instead controls information in active-set contrast directions.
The condition is imposed only over $\mathcal B_t$: a blanket floor would explore every
round, costing $\Omega(T)$ regret under a persistent gap, whereas $\mathcal B_t$
collapses to a singleton once the gap resolves and Eq.~\eqref{eq:coverage} becomes
automatic. Coverage is supplied by the action-selection rule: the
farthest-pair rule attains $\kappa_0=1/4$ (Section~\ref{sec:algorithm}), and
Theorem~\ref{thm:regret} holds for any admissible rule
meeting it. The regret bound itself does not use context diversity
(Assumption~\ref{ass:diversity}): the elliptical-potential sum is algebraic and the
coverage condition supplies the per-round exploration. Context diversity enters only
the identification discussion of Theorem~\ref{thm:variance}(a), through persistent
excitation of every direction.

\begin{assumption}[Offline confidence input]\label{ass:offline}
Independently of the online interaction, the algorithm is supplied a radius
$0\le\bar\eta\le2S$ such that
$\PP(\|\hat\theta^{(0)}-\theta^\star\|_2\le\bar\eta)\ge1-\delta/5$.
After projecting $\hat\theta^{(0)}$ onto the radius-$S$ ball, the deterministic
choice $\bar\eta=2S$ always satisfies this assumption by the triangle inequality.
Any supplied radius can therefore be capped at $2S$.
\end{assumption}

Optionally, any external finite-sample DR procedure supplying a valid radius
$\eta_{\mathrm{off}}$ can be plugged in as $\bar\eta=\eta_{\mathrm{off}}$; a
root-$n_{\mathrm{off}}$ form $\eta_{\mathrm{off}}=\widetilde O(\sqrt{Kd/n_{\mathrm{off}}})$
then yields the warm-start rate below. Such a radius needs procedure-specific
overlap, tail, nuisance, and Gram-matrix conditions in addition to the asymptotic
product-rate condition in Assumption~\ref{ass:nuisance}. The generic choice $2S$
provides a valid default.

\subsection{Regret Decomposition}

\begin{theorem}[Regret]\label{thm:regret}
Under Assumptions~\ref{ass:iid}, \ref{ass:bounded}, \ref{ass:norms},
\ref{ass:realizability}, \ref{ass:subgauss}, \ref{ass:info},
and \ref{ass:offline}, let $D:=Kd$, set
$q_T:=4D\log(9T)+8\log(20T/\delta)$,
$h_T:=27D\log(1+2T/D)+54\log(20T/\delta)$,
and choose $\alpha=4L^2q_T$. Algorithm~\ref{alg:dr-cbwk}, run with
$\epsilon_{\mathrm{floor}}=0$ and the BOSE confidence radius
\begin{equation}
  \beta_t \equiv \beta_T := (\Xi+2F+LS)\sqrt{h_T}
  +\sqrt\alpha\,\bar\eta,
  \label{eq:radius}
\end{equation}
(where $\bar\eta$ is any high-probability upper bound on
$\|\hat\theta^{(0)}-\theta^\star\|_2$: the generic $\bar\eta=2S$ is always
computable, while a certified external radius $\bar\eta=\eta_{\mathrm{off}}$
yields the sharper warm-start term below) and step
size $\eta_t = \eta_0 / \sqrt{T}$ satisfies, for every horizon $T$, with probability at least $1 - \delta$
over the offline sample and the online interaction,
\begin{align}
  R(T) & \leq
  \underbrace{C_1\, D\sqrt{T}\,\log\!\tfrac{KT}{\delta}}_{R_{\mathrm{online}}}
  + \underbrace{C_2 \sqrt{T}}_{R_{\mathrm{dual}}}
  + \underbrace{C_3\,D\sqrt T\,\bar\eta
  \log\!\tfrac{KT}{\delta}}_{R_{\mathrm{warm}}},
  \label{eq:regret}
\end{align}
where $C_1, C_2, C_3$ depend on
$\Xi$, $F$, $R_{\max}$, $c_{\max}$, $c_{\min}$, $\rho$, $L$, $S$,
$\kappa_0, \eta_0$ and
$\log(1/\delta)$ but not on $T$, $n_{\mathrm{off}}$, $d$, or $K$. The constants
grow as the budget rate $\rho$ or $c_{\min}$ shrink, so the guarantee is for a
fixed positive budget rate. Under an external root-$n_{\mathrm{off}}$ guarantee and
the choice $\bar\eta=\eta_{\mathrm{off}}$,
$R_{\mathrm{warm}}=
\widetilde O(D^{3/2}\sqrt{T/n_{\mathrm{off}}})$. With the
generic radius the same overall $\widetilde O(D\sqrt T)$ rate holds. A smaller
certified radius tightens $R_{\mathrm{warm}}$, with $R_{\mathrm{online}}$ and
$R_{\mathrm{dual}}$ retaining their stated rates.
\end{theorem}

The bound is a high-probability frequentist guarantee for the realized
$\theta^\star$. The primal term is the regret of a semiparametric contextual
bandit with the budget offset $-\lambda_t c$, at the rate $\widetilde O(Kd\sqrt T)$
of \citet{krishnamurthy2018semiparametric} for effective dimension $Kd$. The radius
$\beta_T$ comes from the BOSE self-normalized inequality
\citep{krishnamurthy2018semiparametric} rather than the ordinary
predictable-covariate bound, since $u_t$ is realized only with the action.
Equation~\eqref{eq:radius} uses the worst-case residual bound $|f-\hat f|\le2F$,
valid for every baseline estimate clipped to $[-F,F]$. The effect of baseline
accuracy on score variance is characterized in Theorem~\ref{thm:variance}.

\begin{proof}
Let $\mathcal G_t=\sigma(\mathcal H_{t-1},X_t)$ and
$\mathcal F_t=\sigma(\mathcal G_t,A_t,Y_t)$, with the offline inputs included
in $\mathcal H_0$. Set $H:=\Xi+2F+LS$. The residual
$\zeta_t=g_t-u_t^\top\theta^\star$ obeys $|\zeta_t|\le H$ and is conditionally
independent of $u_t$ given $\mathcal G_t$, while $\E[u_t\mid\mathcal G_t]=0$.
The BOSE self-normalized bound (Lemmas 14, 16, and 17 of
\citet{krishnamurthy2018semiparametric}), applied to $u_t/(2L)$ and
$\zeta_t/H$, gives
$\|\sum_{s\le t}u_s\zeta_s\|_{V_t^{-1}}\le H\sqrt{h_T}$ simultaneously
for $t\le T$ with failure probability at most $\delta/5$.
The normal equations and non-expansiveness of the $V_t$-norm projection imply
\[
\|\hat\theta_t-\theta^\star\|_{V_t}
\le H\sqrt{h_T}+\sqrt\alpha\,
\|\hat\theta^{(0)}-\theta^\star\|_2\le\beta_T
\]
on intersection with the offline-radius event. This also holds at $t=0$.

Write $c_t:=c(X_t,A_t)$, $G:=c_{\max}+\rho$, $\eta:=\eta_0/\sqrt T$, and
$\Phi_t(a,\lambda):=\tau_a(X_t)-\lambda(c(X_t,a)-\rho)$.
The projected estimate satisfies $|z_t(a)^\top\hat\theta_{t-1}|\le LS$.
Since $V_{t-1}\succeq\alpha I$, every priced action is eliminated against the
free control when $\lambda_t>(LS+2\beta_TL/\sqrt\alpha)/c_{\min}$.
The queue then decreases; otherwise it can increase by at most $\eta G$.
Induction therefore gives
\begin{equation}
0\le\lambda_t\le\Lambda:=\max\!\left\{\lambda_1,
\frac{LS+2\beta_TL/\sqrt\alpha}{c_{\min}}+\eta G\right\}.
\label{eq:regret-queue-main}
\end{equation}
Under the prescribed parameters, $\beta_T/\sqrt\alpha=O(1)$, so the queue
bound is uniform in the horizon.

For each predictable $\lambda_t\ge0$, weak duality gives
$\OPT_{\mathrm{LP}}(\rho)\le\E_X[\max_a\Phi(a,\lambda_t;X)]$.
The difference between this conditional expectation and
$\max_a\Phi_t(a,\lambda_t)$ is a martingale difference, bounded by
$2(R_{\max}+\Lambda G)$. Conditional on the offline inputs, the queue bound
holds on every path. Azuma-Hoeffding therefore gives, with probability at least
$1-\delta/5$,
\begin{equation}
\begin{aligned}
R(T)\le{}&\sum_{t=1}^T[\max_{a\in\A}\Phi_t(a,\lambda_t)-\Phi_t(A_t,\lambda_t)]\\
&+\sum_{t=1}^T\lambda_t(\rho-c_t)
+O(\sqrt{T\log(5/\delta)}).
\end{aligned}
\label{eq:regret-master-main}
\end{equation}
The dual term is controlled deterministically: squaring
$\lambda_{t+1}=[\lambda_t+\eta(c_t-\rho)]_+$ and telescoping yields
\begin{equation}
\sum_{t=1}^T\lambda_t(\rho-c_t)
\le\frac{\lambda_1^2}{2\eta}+\frac{\eta TG^2}{2}=O(\sqrt T).
\label{eq:regret-dual-main}
\end{equation}

To bound the primal term, let
$a_t^\star\in\argmax_{a\in\A_t^{\mathrm{aff}}}\Phi_t(a,\lambda_t)$.
On the confidence event, its estimated score disadvantage against any
affordable action is at most one confidence width. It therefore survives the
two-width elimination test, so both $a_t^\star$ and $A_t$ belong to $\mathcal B_t$.
The active-set test bounds their estimated score gap by twice the confidence
width; the estimation error contributes one more width. Write
$\ell_t:=\Phi_t(a_t^\star,\lambda_t)-\Phi_t(A_t,\lambda_t)$.
Defining
$w_t:=\E[\|u_t\|_{V_{t-1}^{-1}}^2\mid X_t,\mathcal H_{t-1}]$, coverage gives
\begin{equation}
\begin{aligned}
\ell_t
&\le3\beta_T\|z_t(a_t^\star)-z_t(A_t)\|_{V_{t-1}^{-1}}\\
&\le3\beta_T\sqrt{w_t/\kappa_0}.
\end{aligned}
\label{eq:regret-gap-main}
\end{equation}
For $\Delta_t:=\|u_t\|_{V_{t-1}^{-1}}^2\le b:=4L^2/\alpha\le1$,
$\E[\Delta_t\mid\mathcal G_t]=w_t$. Since $e^{-x}\le1-x/2$ on $[0,1]$,
\[
\E[e^{-\Delta_t/b}\mid\mathcal G_t]
\le1-w_t/(2b)\le e^{-w_t/(2b)}.
\]
The tower property makes
$\exp\{\sum_{t\le s}(w_t/(2b)-\Delta_t/b)\}$ a nonnegative
$\mathcal F_s$-supermartingale starting at one. Markov's inequality gives,
with probability at least $1-\delta/5$,
\begin{equation}
\sum_{t=1}^T w_t\le2\sum_{t=1}^T\Delta_t+2b\log(5/\delta).
\label{eq:potential-transfer}
\end{equation}
The determinant identity, $\Delta_t\le1$, and the trace bound yield
$\sum_t\Delta_t\le2\log(\det V_T/\det V_0)
\le2D\log(1+4L^2T/(\alpha D))$ \citep{abbasi2011improved}.
Thus $\sum_t w_t\le\mathcal P_T$, where
$\mathcal P_T:=4D\log(1+4L^2T/(\alpha D))+2b\log(5/\delta)$.
Equation~\eqref{eq:regret-gap-main} and Cauchy-Schwarz give
\begin{equation}
\sum_{t=1}^T\ell_t\le3\beta_T\sqrt{T\mathcal P_T/\kappa_0}.
\label{eq:ts-ideal}
\end{equation}
The stochastic component of $\beta_T$ contributes $\widetilde O(D\sqrt T)$;
the prior component contributes
\begin{equation}
\begin{aligned}
R_{\mathrm{warm}}&\le3\sqrt\alpha\,\bar\eta\sqrt{T\mathcal P_T/\kappa_0}\\
&=\widetilde O(D\sqrt T\,\bar\eta).
\end{aligned}
\label{eq:warm}
\end{equation}
The definitions of $h_T$, $q_T$, and $\mathcal P_T$ give the explicit
$D\sqrt T\log(KT/\delta)$ factors in Eq.~\eqref{eq:regret}. An external
$\bar\eta=\widetilde O(\sqrt{D/n_{\mathrm{off}}})$ gives the stated
root-$n_{\mathrm{off}}$ warm-start rate.

Finally, the queue update also implies
$S_t\le t\rho+(\lambda_{t+1}-\lambda_1)/\eta\le t\rho+\Lambda/\eta$.
Thus every action is affordable except possibly during the last
$\lceil\Lambda/(\eta\rho)+c_{\max}/\rho\rceil=O(\sqrt T)$ rounds.
The global and affordable maxima of $\Phi_t$ differ by at most $R_{\max}$:
the free control is always affordable and each priced action has
$\tau_a-\lambda_t c\le R_{\max}$. Affordability therefore adds only
$O(R_{\max}\sqrt T)$ to Eq.~\eqref{eq:regret-master-main} while enforcing
$S_t\le B$ on every path. A union bound over the offline-radius, confidence,
context-concentration, and potential-concentration events, each with failure
probability at most $\delta/5$, proves Eq.~\eqref{eq:regret} with probability
at least $1-\delta$.
\end{proof}

\subsection{Baseline Robustness of Action-Centering}

The action-centered update targets the treatment contrasts $\tau_a$ through the
score $s_t^{\hat f}:=u_t\,(Y_t-\hat f(X_t))$. We show this is exact for an
arbitrary baseline, characterize the bias of a specified separate-arm linear
comparator, and identify when oracle residualization reduces total score variance.
Fix a round $t$ and condition on $(X_t,\mathcal H_{t-1})$; write
$r_t:=f(X_t)-\hat f(X_t)$ for the predictable baseline residual (so $\hat f\equiv0$
gives $r_t=f(X_t)$, the unresidualized centered score). Write
$\tau_a:=\tau_a(X_t)$ and
$\sigma^2(x,a):=\Var(Y_t\mid X_t=x,A_t=a,\mathcal H_{t-1})$;
expectations and covariances below condition on $(X_t,\mathcal H_{t-1})$ unless
specified otherwise.

\begin{theorem}[Baseline robustness]\label{thm:variance}
Under Assumptions~\ref{ass:iid}, \ref{ass:bounded}, \ref{ass:norms},
\ref{ass:realizability}, and~\ref{ass:subgauss},
the following hold. Only the confidence-event bound in~(a) and the regret theorem
invoke the action-independence of Assumption~\ref{ass:subgauss}; the moment identity
in~(a) and the covariance decomposition in~(c) require only conditionally mean-zero
bounded noise and hold for a general action-dependent conditional variance
$\sigma^2(x,a)$.

(a) For every predictable $\hat f$,
$\E[s_t^{\hat f}\mid X_t,\mathcal H_{t-1}]=M_t\theta^\star$, with no
inverse-propensity weighting. For $|\hat f|\le F$ and under the confidence conditions of
Theorem~\ref{thm:regret}, with probability at least $1-\delta$,
$|v_t^\top(\hat\theta_t-\theta^\star)|\le\beta_t\|v_t\|_{V_t^{-1}}$ for every
direction $v_t$. Directional accuracy is governed by the confidence width
$\beta_t\|v_t\|_{V_t^{-1}}$. Context diversity together with persistent randomization spanning
every arm block can supply full-parameter identification; neither is required
after elimination resolves a contrast. When the active law is a singleton,
$M_t=0$ and that round adds no contrast information.

(b) For an i.i.d. logged sample, consider the separate-arm population OLS
coefficients
\begin{equation}
b_a^{\mathrm{raw}}:=\argmin_b\E[(Y-X^\top b)^2\mid A=a],
G_a:=\E[XX^\top\mid A=a]\succ0.
\label{eq:raw-comparator}
\end{equation}
Let $\gamma_a:=G_a^{-1}\E[Xf(X)\mid A=a]$ be the coefficient of the
arm-specific linear projection of $f$. Then
$b_a^{\mathrm{raw}}=\theta_a^\star+\gamma_a$ and the fitted contrast against
control has pointwise bias $x^\top(\gamma_a-\gamma_{a_0})$. Under the usual
finite-second-moment conditions, empirical separate-arm OLS converges to these
coefficients as every arm count diverges. Thus its asymptotic contrast bias in
$L^2(P_X)$ is exactly
$\|x^\top(\gamma_a-\gamma_{a_0})\|_{L^2(P_X)}$, which is zero under
context-independent assignment but can be nonzero under action-dependent context
sampling and a misspecified linear reward model. For comparison, by
Eq.~\eqref{eq:policy-equivalence}, a correctly specified reward learner targets
the same constrained optimum as the CATE learner.

(c) The score covariance splits as
\begin{equation}
\begin{aligned}
&\mathrm{Cov}(s_t^{\hat f}\mid X_t,\mathcal H_{t-1}) =\\
&\quad \underbrace{\E[u_tu_t^\top\sigma^2(X_t,A_t)]}_{\text{irreducible noise}}\\
&\quad + \underbrace{r_t^2 M_t + 2 r_t\,\E[u_tu_t^\top\tau_{A_t}]
+ \mathrm{Cov}(u_t\tau_{A_t})}_{\text{signal design}},
\end{aligned}
\label{eq:variance-decomp}
\end{equation}
where only the terms carrying $r_t$ depend on $\hat f$. Comparing the unresidualized score
($\hat f\equiv0$, $r_t=f(X_t)$) with the oracle-residualized learner
($\hat f=f$, $r_t=0$),
\begin{equation}
\mathrm{tr}\,\mathrm{Cov}(s_t^{0}) - \mathrm{tr}\,\mathrm{Cov}(s_t^{f})
= f(X_t)^2\,\mathrm{tr}(M_t) + 2 f(X_t)\,\E[\|u_t\|_2^2\,\tau_{A_t}],
\label{eq:variance-trace}
\end{equation}
which is strictly positive whenever $\mathrm{tr}(M_t)>0$
and $|f(X_t)| > 2\,|\E[\|u_t\|_2^2\tau_{A_t}]|/\mathrm{tr}(M_t)$, in particular
whenever the baseline scale exceeds twice the treatment-effect scale $R_{\max}$;
when the active set is a singleton, $M_t=0$ and both score covariances vanish.
\end{theorem}

Part (a) is the semiparametric
action-centering identity
\citep{greenewald2017action,krishnamurthy2018semiparametric}: because
$\E[u_t\mid X_t,\mathcal H_{t-1}]=0$, the
baseline and any error in its estimate cancel in the mean, preserving the exact
moment under a nonlinear or misspecified baseline. Part (b) is the
complementary calculation obtained by substituting the outcome model into the
OLS normal equations. Part (c) follows from total covariance conditional on the
action and quantifies the oracle reduction in total score variance: exact
residualization removes the $O(f^2)$ inflation once the baseline scale dominates
the treatment-effect scale. An estimated residualizer always preserves the
estimating moment but can increase or decrease variance depending on its error;
clipping supplies the valid $2F$ residual bound in Eq.~\eqref{eq:radius}.

\section{Experiments}
\label{sec:experiments}

We evaluate DR-CATE-CBwK using real job-promotion features and simulated outcomes,
treatment effects, logging actions, and offline-to-online shift. Known simulator
CATEs and the fluid LP optimum allow direct regret evaluation without estimating
counterfactual outcomes from a held-out observational log. We examine
operating-regime value, budget feasibility, finite-horizon scaling, baseline
robustness, and nonlinear CATEs.

\subsection{Experiment Setup}
\label{sec:exp-setup}

Contexts are drawn from proprietary standardized promotion-log features ($d=8$
continuous job, company, and poster attributes). There are five total actions: a
no-offer control $a_0$ and $K=4$ priced promotions (a free trial and \$50, \$100,
\$200 credits), so the stacked parameter has effective dimension $Kd=32$.
The effective per-decision costs are \$15, \$30, \$55, and \$90, respectively,
incorporating empirical redemption rates and normalized by the largest cost.
The budget is enforced on this normalized effective spend, a proxy for cash
liability; a guaranteed cash cap requires a conservative cost bound for each
promotion. We impose the
linear CATE $\tau_a(x)=x^\top\theta_a^\star$ (Assumption~\ref{ass:realizability})
over an action-independent baseline $f$. Outcomes follow
$Y=f(X)+\tau_A(X)+\xi$, with bounded noise
$\xi\sim\mathrm{clip}(\cN(0,\sigma^2))$ (Assumption~\ref{ass:subgauss}); unless noted,
$f$ is nonlinear, using a tanh-and-interaction surface.
The offline log of $n_{\mathrm{off}}=10^4$ rows is generated by a
context-dependent $\epsilon$-greedy logging policy ($\epsilon=0.15$) on a noisy
budget-adjusted CATE proxy, so the logged action depends on the context (the
action-dependent-assignment regime of Theorem~\ref{thm:variance}(b), consistent with
conditional unconfoundedness); its exact propensities are recorded. An
offline-to-online shift parameter in $[0,1]$ makes the offline log's arm
responsiveness differ from the online one; the online environment uses the current,
fixed CATE $\theta^\star$, against which $\OPT_{\mathrm{LP}}(\rho)$ and regret are
computed exactly. The online horizon is
$T=6000$ at budget rate
$\rho=0.10$; we report means over $32$ seeds with $95\%$ confidence intervals, and
assess close comparisons with paired-seed bootstrap tests.

The offline warm start is Algorithm~\ref{alg:offline-dr}: gradient-boosted
cross-fit nuisances form the DR pseudo-outcome, a ridge regression of which yields
$\hat\theta^{(0)}$, and $\hat f=\hat\mu(\cdot,a_0)$ is the baseline model; the
online estimator seeds its prior mean at $\hat\theta^{(0)}$ with $V_0=\alpha I$. The
online action rule is the practical instantiation of Section~\ref{sec:algorithm}:
active-set elimination (Eq.~\eqref{eq:active-set}) followed by Thompson-type
randomization restricted to $\mathcal B_t$, with the virtual-queue dual update
(Eq.~\eqref{eq:dual-update}) at step $\eta_t=\eta_0/\sqrt T$. We form $\pi_t$ as the
explicit empirical categorical law over $\mathcal B_t$ induced by a fixed set of
posterior draws, so the sampling law, and hence $\bar z_t$, is known exactly. We
compare this empirical Thompson-restricted rule with farthest-pair and
coverage-mixture laws. The latter satisfy Assumption~\ref{ass:info}, whereas
Thompson sampling need not. Theorem~\ref{thm:regret} analyzes coverage-based
selection with prescribed confidence parameters; the experiments use calibrated
confidence and dual scales. The affordable-set cap is unchanged across variants.

\subsection{Baselines and Ablations}
\label{sec:exp-methods}

LinUCB \citep{li2010contextual} fits linear raw rewards, with an affordable-set cap
for feasibility. UCB-ALP \citep{wu2015algorithms} combines per-arm reward UCBs with
fluid-LP pacing. SquareCBwK \citep{slivkins2023contextual}
uses a reward-regression oracle, inverse-gap exploration, and primal-dual control.

Raw-CBwK uses our controller and affordable set with separate-arm linear reward
regression, targeting $Q_a-Q_{a_0}$ where $Q_a(x)=\E[Y\mid X=x,A=a]$.
Raw-Flex-CBwK additionally uses our offline baseline $\hat f$, fitting separate-arm
ridge heads to $Y-\hat f$; this matches baseline-model capacity but retains
uncentered regression. Offline-DR freezes the offline policy with a budget-tuned
dual and no online learning.

The No CATE prior ablation sets $\hat\theta^{(0)}=0$ while retaining $\hat f$;
No residualization sets $\hat f=0$ while retaining the CATE prior. These isolate
the prior and residualization components, respectively. The former is the closest
BOSE-style learner under our controller
\citep{krishnamurthy2018semiparametric}. No budget removes both the dual update
and the affordable-set restriction.

\subsection{Operating Regime: Value and Regret}
\label{sec:exp-operating}

The operating regime combines a nonlinear baseline, context-dependent logging,
and an offline-to-online shift (shift $=1$).
Table~\ref{tab:operating} reports regret against the fluid LP and total spend. Among
all budget-feasible methods DR-CATE-CBwK has the lowest regret ($498$), ahead of the
zero-CATE-prior variant, the capacity-matched Raw-Flex-CBwK, the CBwK and reward
baselines, and the frozen offline policy whose stale ranking mis-targets the drifted
effects. Paired-seed bootstrap comparisons over the $32$ shared seeds show that the CATE
prior lowers regret by $94$
($95\%$ CI $[29,160]$, $p{=}0.006$) over the zero-CATE-prior variant (with offline
residualization retained); residualization by $338$ ($[237,445]$, $p{<}0.001$); and
DR-CATE-CBwK improves by $220$ over Raw-CBwK and $430$ over SquareCBwK (both
$p{<}0.001$). Conditional reward and CATE share the same constrained optimizer
(Eq.~\eqref{eq:policy-equivalence}). A correctly specified reward learner makes the
same budget-adjusted comparison, since the shared $f(X_t)$ cancels from
$Q_a-\lambda c$. The gap instead reflects these learners' exposure to shared-baseline
misspecification and to action-dependent context distributions across separate arm
models, together with differences in exploration and budget control. The
capacity-matched Raw-Flex-CBwK improves on Raw-CBwK by $98$ ($[54,144]$,
$p{<}0.001$), a baseline-capacity gap, yet DR-CATE-CBwK still beats it by $122$
($[60,178]$, $p{<}0.001$), so action-centering's robustness survives capacity
matching. The only method with
lower regret, the no-budget ablation, reaches it only by spending $2729$ against the
$600$ cap, about $4.5$ times the budget. The affordable-set restriction enforces
feasibility. With the same seeds we evaluate the farthest-pair law and the coverage
mixture $\pi_t=(1-\gamma)q_t^{\mathrm{TS}}+\gamma\pi_t^{\mathrm{FP}}$
($\gamma{=}0.1$, $\kappa_0=\gamma/4$; Section~\ref{sec:algorithm}). Their regrets
are $1285$ and $822$, respectively, versus $498$ for Thompson, and both respect
the budget. The mixture improves on farthest-pair while retaining active-set
coverage, and Thompson achieves the lowest regret in this comparison.
DR-CATE-CBwK spends $599.9$ of the $600$ cap. Its edge over the frozen offline policy is
driven by the shift: as the shift sweeps $[0,1]$ the static policy's regret
rises from $257$ to approximately $1200$ while ours stays near $500$ (crossing near $0.4$), and
Raw-CBwK is flat but uniformly worse under context-dependent logging.

\begin{table}[!htp]
  \centering
  \small
  \caption{Operating regime (nonlinear baseline, context-dependent logging,
  offline-to-online shift; $T{=}6000$, $\rho{=}0.10$, $32$ seeds, $95\%$ CI). Regret is
  against the fluid LP (lower is better); $\pm$ is the $95\%$ CI half-width over
  seeds. DR-CATE-CBwK variants use calibrated parameters as described in
  Section~\ref{sec:exp-setup}. Spend is rounded to whole normalized effective-cost
  units. Budget-aware methods respect the $600$ cap;
  the No budget ablation spends $2729$.}
  \label{tab:operating}
  \begin{tabular}{lrr}
    \toprule
    Method & Regret $R(T)$ & Spend ($B{=}600$) \\
    \midrule
    DR-CATE-CBwK (Thompson, ours)   & $\mathbf{498}{\scriptstyle\pm56}$ & $600$ \\
    No CATE prior         & $592{\scriptstyle\pm54}$  & $600$ \\
    Raw-Flex-CBwK         & $620{\scriptstyle\pm38}$  & $600$ \\
    Raw-CBwK              & $718{\scriptstyle\pm54}$  & $600$ \\
    No residualization    & $836{\scriptstyle\pm98}$ & $600$ \\
    SquareCBwK            & $928{\scriptstyle\pm103}$ & $600$ \\
    UCB-ALP               & $998{\scriptstyle\pm80}$ & $594$ \\
    LinUCB                & $1040{\scriptstyle\pm77}$ & $600$ \\
    Offline-DR (fixed)    & $1214{\scriptstyle\pm128}$ & $582$ \\
    \midrule
    DR-CATE-CBwK (mixture)       & $822{\scriptstyle\pm48}$ & $600$ \\
    DR-CATE-CBwK (farthest-pair) & $1285{\scriptstyle\pm87}$ & $600$ \\
    \midrule
    No budget (infeasible) & $-130{\scriptstyle\pm91}$ & $2729$ \\
    \bottomrule
  \end{tabular}
\end{table}

\subsection{Regret Rate}
\label{sec:exp-rate}

\begin{figure}[!htp]
  \centering
  \includegraphics[width=0.6\linewidth]{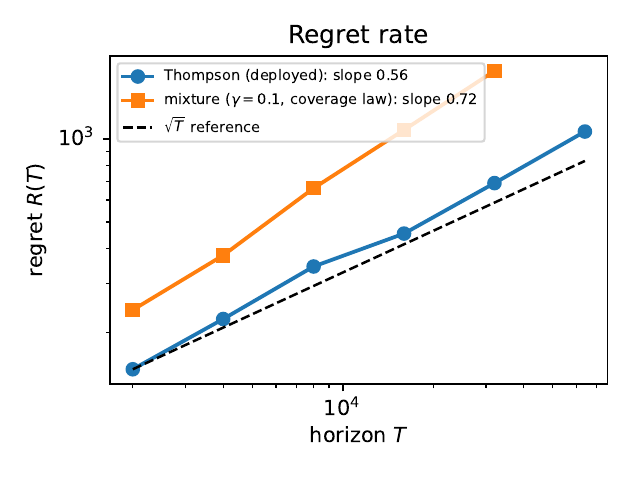}
  \caption{Finite-horizon scaling for Thompson (deployed; slope $0.56$, CI $[0.51,0.61]$)
  and the coverage mixture ($\gamma{=}0.1$, slope $0.72$), using calibrated
  confidence and dual scales.}
  \label{fig:ac-rate}
\end{figure}

To examine finite-horizon scaling, we remove the offline-to-online shift, relax
the budget, and use a linear baseline with the fitted offline $\hat f$.
Figure~\ref{fig:ac-rate} plots $\log R(T)$ against $\log T$. The practical Thompson
rule, labeled ``Thompson (deployed)'' in the figure, has a fitted exponent
of $0.56$ up to $6.4\times10^4$ rounds (bootstrap CI
$[0.51,0.61]$). The mixture and farthest-pair slopes up to $3.2\times10^4$ rounds
are $0.72$ ($[0.69,0.76]$) and $0.85$. These estimates describe finite-horizon
scaling under the calibrated settings, with Thompson showing the slowest regret
growth over the evaluated ranges.

\subsection{Baseline Robustness}
\label{sec:exp-robust}

\begin{figure}[!htp]
  \centering
  \includegraphics[width=0.6\linewidth]{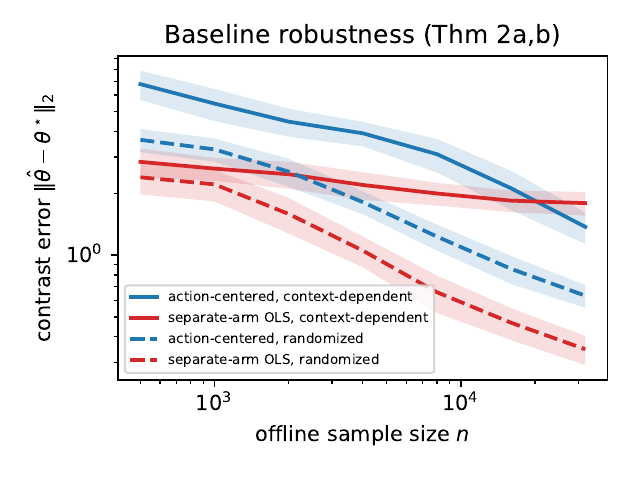}
  \caption{Baseline robustness (Theorem~\ref{thm:variance}(a,b)). Contrast error versus
  sample size. Under context-dependent logging the action-centered estimator's error
  keeps decreasing while separate-arm OLS plateaus at a bias; under randomized logging
  both decrease. Means over $32$ seeds.}
  \label{fig:ac-est}
\end{figure}

We examine the predictions of Theorem~\ref{thm:variance} on the estimator, separately from
the bandit loop. Figure~\ref{fig:ac-est} measures the contrast error
$\|\hat\theta-\theta^\star\|_2$ as the logged sample grows. Under context-dependent logging
the action-centered estimator (run with $\hat f\equiv0$, so centering alone removes
the baseline) has error decreasing throughout, while the separate-arm
OLS comparator plateaus at a bias floor, consistent with
Theorem~\ref{thm:variance}(a,b) over the evaluated sample sizes.
Under randomized logging both estimators' errors decrease, consistent with the
role of context-dependent assignment in the comparator's bias. Varying the baseline
magnitude at fixed sample size gives approximately constant error for the residualized
estimator and linearly increasing error for the unresidualized centered estimator
($\hat f\equiv0$), consistent with Theorem~\ref{thm:variance}(c). Residualizer
accuracy also matters:
with $|f|\!\approx\!4R_{\max}$ the contrast error is $0.17$ under an oracle $\hat f$
and $3.2$ with none, but a badly misspecified $\hat f$ raises it to $8.1$, above the
unresidualized estimator, illustrating how residualization quality affects variance.

\subsection{Coverage, Sensitivity, and Calibration}
\label{sec:exp-coverage}

We measure the coverage condition for the empirical action rules
(Assumption~\ref{ass:info}) empirically by logging, per multi-action round, the
achieved constant
$\kappa_t:=w_t/\max_{a\ne a'\in\mathcal B_t}\|z_t(a)-z_t(a')\|_{V_{t-1}^{-1}}^2$
(zero-distance pairs excluded), where
$w_t=\E_{A\sim\pi_t}[\|z_t(A)-\bar z_t\|_{V_{t-1}^{-1}}^2\mid X_t,\mathcal H_{t-1}]$.
This is the largest constant certified by Eq.~\eqref{eq:coverage}. The
farthest-pair rule attains $\kappa_0=1/4$ in each evaluated multi-action round.
The Thompson rule has median coverage $0$ and mean $0.02$ over
$\approx4.7\times10^3$ such rounds per seed, as it concentrates draws on the
apparent best arm. It achieves
lower regret (Table~\ref{tab:operating}), illustrating the gap between sufficient
coverage and the empirical behavior of Thompson sampling. The affordable-set
restriction preserves budget feasibility for both rules.

The raw offline error grows with drift ($\|\hat\theta^{(0)}_{\mathrm{raw}}-\theta^\star\|\approx1.53$
stationary, $2.63$ under full drift). With $S=2$ (Assumption~\ref{ass:norms};
$\|\theta^\star\|\approx0.8$), projection onto the radius-$S$ ball leaves
post-projection errors $\approx1.5$ and $2.0$, within $\bar\eta=2S=4$. The larger
initial error under shift motivates online adaptation. The empirical
implementation uses calibrated confidence, Thompson covariance, and dual scales
rather than the theorem's bound-based settings. Scales are fixed across seeds.
A post-hoc $27$-point sensitivity grid gives regret in $[345,773]$ (median $537$).
Scoring against the known simulator optimum is oracle-assisted; a practical
calibration would use an estimable validation objective.

\subsection{Robustness to the Linear-CATE Assumption: a Neural Variant}
\label{sec:exp-nonlinear}

To examine CATE misspecification (Assumption~\ref{ass:realizability}), we
replace the linear CATE by a nonlinear, non-monotone surface
($\tau_a(x)=\kappa[\sin(\omega\,x^\top u_a)+\sin(\omega\,x^\top v_a+1)]$, with amplitude
$\kappa$, frequency $\omega=5$, and fixed random per-arm directions $u_a,v_a$)
that a linear head cannot represent, keeping everything else fixed. The neural
variant (Neural DR-CATE-CBwK) runs the action-centered algorithm in a learned
feature space: the offline stage fits a one-hidden-layer $\tanh$ network to the
cross-fitted contrasts and uses its representation $\phi(x)$ in place of $x$
($e_a\otimes\phi(X_t)$), leaving everything causal and the budget controller
unchanged. This provides an empirical extension to learned representations.

Table~\ref{tab:nonlinear} reports the result. Under a nonlinear CATE the linear
DR-CATE-CBwK has regret $945$, lower than every
reward-based baseline (Table~\ref{tab:nonlinear}), with the margin over the
capacity-matched Raw-Flex-CBwK confirmed by a paired test ($64$, $[28,100]$,
$p{=}0.001$). Action-centering still cancels the shared baseline in the moment.
The neural variant has regret $906$; its improvement over the linear head ($39$, paired
$p{=}0.11$) is not statistically significant. Both variants retain affordable-set
budget feasibility.

\begin{table}[!htp]
  \centering
  \small
  \caption{Nonlinear CATEs ($\omega=5$, $T{=}6000$, $\rho{=}0.10$, $32$ seeds;
  $\pm$ denotes $95\%$ CI half-width). The neural improvement over the linear
  variant is not significant ($p{=}0.11$).}
  \label{tab:nonlinear}
  \begin{tabular}{lr}
    \toprule
    Method & Regret $R(T)$ \\
    \midrule
    Neural DR-CATE-CBwK   & $\mathbf{906}{\scriptstyle\pm42}$ \\
    DR-CATE-CBwK (linear) & $945{\scriptstyle\pm45}$ \\
    Raw-Flex-CBwK         & $1009{\scriptstyle\pm45}$ \\
    Raw-CBwK              & $1063{\scriptstyle\pm46}$ \\
    UCB-ALP               & $1070{\scriptstyle\pm50}$ \\
    SquareCBwK            & $1078{\scriptstyle\pm57}$ \\
    LinUCB                & $1088{\scriptstyle\pm50}$ \\
    \bottomrule
  \end{tabular}
\end{table}

\section{Conclusion}
\label{sec:conclusion}

We presented DR-CATE-CBwK, an online causal contextual bandit combining a doubly
robust CATE warm start, BOSE-style action-centering, and a global budget. The joint
design couples an offline causal prior and semiparametric active-set learning with
a primal-dual budget controller. It therefore avoids modeling the arbitrary baseline
online and does not require permanent exploration. For linear CATEs in the stated
stationary i.i.d. setting, Theorem~\ref{thm:regret} gives a high-probability
$\widetilde O(Kd\sqrt{T})$ bound against the fluid LP for the coverage-guaranteed
action rule. Theorem~\ref{thm:variance} establishes baseline-free centered moments,
characterizes when a specified separate-arm comparator can be biased, and
identifies when oracle residualization reduces score variance.
On a semi-synthetic benchmark based on an online job-promotion targeting dataset,
the practical Thompson-restricted variant improves on a raw-reward learner and a
static offline policy under an offline-to-online shift and shows sublinear
finite-horizon scaling, while the separately evaluated farthest-pair and
coverage-mixture variants are more conservative. The centered estimator recovers
contrasts where the specified uncentered learner is biased, and all budget-aware
variants respect the effective-cost budget (Section~\ref{sec:experiments}).
Action-centering supplies the statistical robustness, the offline prior improves
adaptation to an offline-to-online shift, and the affordable-set primal-dual controller
supplies pathwise feasibility. Together, these components turn incremental-effect
estimation into an online, budget-aware decision process while preserving the
distinction between the coverage-guaranteed action laws and the stronger empirical
Thompson variant.

Each structural choice suggests a natural extension: continuous doses via kernel
smoothing, non-i.i.d.\ data via discounted updates, and multiple resources via a
multi-dimensional dual.
\ifincludeproofs
 \ifincludeneuralntk
The supplement adds frozen-representation and online NTK extensions under
explicit representation-quality and fixed-NTK coupling conditions.
 \else
The supplement extends the analysis beyond linear CATEs through a frozen neural
representation. Its regret theorem conditions on certified representation
quality; deriving that guarantee from a specific training procedure and analyzing
online representation updates remain open.
 \fi
\fi
Our guarantees assume a fixed positive budget rate, motivating further study of
residual-budget-adaptive exploration and the optimal small-budget rate.
Public uplift datasets and live traffic provide natural next steps for validation.

\section*{Ethical Considerations}
Promotion targeting allocates financial benefits across users. Equitable
allocation requires attention to historical disparities:
underserved groups may have noisier effect estimates and receive fewer offers,
further reducing opportunities to learn their responses. Deployment should include
allocation and outcome audits across protected and vulnerable groups, human
oversight, and recourse alongside the financial budget.

Offline causal estimates depend on unconfoundedness given the recorded features,
so logging validation should assess potential unobserved confounding. Online
randomization supports identification, but eliminated actions may receive no
further observations. Randomized holdouts with suitable overlap and monitoring of
predicted versus observed effects are needed to detect persistent errors and
feedback effects. Valid evaluation of an alternative policy requires overlap
between its actions and those represented in the evaluation data.

Behavioral features and propensities require data minimization, access controls,
and retention limits. Sensitive attributes should be used only when lawful and
necessary, including for audited fairness assessments. Operators should protect
logs and online updates against poisoning, feature manipulation, and unauthorized
access. Appropriate offer restrictions, exposure limits, and human review are
needed to prevent manipulative targeting. The affordable-set rule caps the
specified effective costs; a cash guarantee additionally requires conservative
cost accounting for variation in redemption.

Our experiments use de-identified logged features and simulated outcomes.
For deployment, we recommend causal validation, privacy review, security controls,
and assessment of distributional harms.

\bibliographystyle{unsrtnat}
\bibliography{references}

@inproceedings{greenewald2017action,
  author    = {Greenewald, Kristjan and Tewari, Ambuj and Klasnja, Predrag and Murphy, Susan},
  title     = {Action Centered Contextual Bandits},
  booktitle = {Advances in Neural Information Processing Systems},
  volume    = {30},
  publisher = {Curran Associates, Inc.},
  address   = {Red Hook, NY, USA},
  numpages  = {9},
  year      = {2017},
  url       = {https://arxiv.org/abs/1711.03596}
}

@inproceedings{zhang2024disco,
  author    = {Zhang, Jason Shuo and Howson, Benjamin and Savva, Panayiota and Loh, Eleanor},
  title     = {DISCO: An End-to-End Bandit Framework for Personalised Discount Allocation},
  booktitle = {Machine Learning and Knowledge Discovery in Databases: Applied Data Science Track (ECML PKDD)},
  series    = {Lecture Notes in Computer Science},
  volume    = {14950},
  pages     = {33--49},
  publisher = {Springer},
  address   = {Vilnius, Lithuania},
  year      = {2024}
}

@misc{sawant2018contextual,
  author        = {Sawant, Neela and Namballa, Chitti Babu and Sadagopan, Narayanan and Nassif, Houssam},
  title         = {Contextual Multi-Armed Bandits for Causal Marketing},
  year          = {2018},
  eprint        = {1810.01859},
  archivePrefix = {arXiv},
  primaryClass  = {cs.LG},
  note          = {ICML 2018 Workshops}
}

@inproceedings{zhao2019uplift,
  author    = {Zhao, Zhenyu and Harinen, Totte},
  title     = {Uplift Modeling for Multiple Treatments with Cost Optimization},
  booktitle = {2019 IEEE International Conference on Data Science and Advanced Analytics (DSAA)},
  pages     = {422--431},
  publisher = {IEEE},
  address   = {Washington, DC, USA},
  year      = {2019}
}

@inproceedings{kim2025experimental,
  author    = {Kim, Seok-Jin and Kim, Gi-Soo and Oh, Min-hwan},
  title     = {Experimental Design for Semiparametric Bandits},
  booktitle = {Conference on Learning Theory (COLT)},
  series    = {Proceedings of Machine Learning Research},
  volume    = {291},
  pages     = {3215--3252},
  publisher = {PMLR},
  address   = {Lyon, France},
  year      = {2025}
}

@inproceedings{agrawal2016efficient,
  author    = {Agrawal, Shipra and Devanur, Nikhil R. and Li, Lihong},
  title     = {An Efficient Algorithm for Contextual Bandits with Knapsacks, and an Extension to Concave Objectives},
  booktitle = {Conference on Learning Theory (COLT)},
  series    = {Proceedings of Machine Learning Research},
  volume    = {49},
  pages     = {4--18},
  publisher = {PMLR},
  address   = {New York, NY, USA},
  year      = {2016}
}

@inproceedings{krishnamurthy2018semiparametric,
  author    = {Krishnamurthy, Akshay and Wu, Zhiwei Steven and Syrgkanis, Vasilis},
  title     = {Semiparametric Contextual Bandits},
  booktitle = {Proceedings of the 35th International Conference on Machine Learning (ICML)},
  series    = {Proceedings of Machine Learning Research},
  volume    = {80},
  pages     = {2776--2785},
  publisher = {PMLR},
  address = {Stockholm, Sweden},
  year      = {2018}
}

@article{athey2021policy,
  author  = {Athey, Susan and Wager, Stefan},
  title   = {Policy Learning with Observational Data},
  journal = {Econometrica},
  volume  = {89},
  number  = {1},
  pages   = {133--161},
  year    = {2021}
}

@article{badanidiyuru2018bandits,
  author  = {Badanidiyuru, Ashwinkumar and Kleinberg, Robert and Slivkins, Aleksandrs},
  title   = {Bandits with Knapsacks},
  journal = {Journal of the ACM},
  volume  = {65},
  number  = {3},
  pages   = {13:1--13:55},
  year    = {2018}
}

@article{balseiro2019learning,
  author  = {Balseiro, Santiago and Gur, Yonatan},
  title   = {Learning in Repeated Auctions with Budgets: Regret Minimization and Equilibrium},
  journal = {Management Science},
  volume  = {65},
  number  = {9},
  pages   = {3952--3968},
  year    = {2019}
}

@inproceedings{bibaut2021post,
  author    = {Bibaut, Aur{\'e}lien and Dimakopoulou, Maria and Kallus, Nathan and Chambaz, Antoine and van der Laan, Mark},
  title     = {Post-Contextual-Bandit Inference},
  booktitle = {Advances in Neural Information Processing Systems (NeurIPS)},
  pages     = {28548--28559},
  publisher = {Curran Associates Inc.},
  address   = {Red Hook, NY, USA},
  year      = {2021}
}

@article{dudik2014doubly,
  author  = {Dud{\'i}k, Miroslav and Erhan, Dumitru and Langford, John and Li, Lihong},
  title   = {Doubly Robust Policy Evaluation and Optimization},
  journal = {Statistical Science},
  volume  = {29},
  number  = {4},
  pages   = {485--511},
  year    = {2014}
}

@article{foster2023orthogonal,
  author  = {Foster, Dylan J. and Syrgkanis, Vasilis},
  title   = {Orthogonal Statistical Learning},
  journal = {The Annals of Statistics},
  volume  = {51},
  number  = {3},
  pages   = {879--908},
  year    = {2023}
}

@article{kennedy2023towards,
  author  = {Kennedy, Edward H.},
  title   = {Towards Optimal Doubly Robust Estimation of Heterogeneous Causal Effects},
  journal = {Electronic Journal of Statistics},
  volume  = {17},
  number  = {2},
  pages   = {3008--3049},
  year    = {2023}
}

@misc{kim2021doubly,
  author        = {Kim, Wonyoung and Kim, Gi-Soo and Paik, Myunghee Cho},
  title         = {Doubly Robust Thompson Sampling for Linear Payoffs},
  year          = {2021},
  eprint        = {2102.01229},
  archivePrefix = {arXiv},
  primaryClass  = {stat.ML}
}

@article{kunzel2019metalearners,
  author  = {K{\"u}nzel, S{\"o}ren R. and Sekhon, Jasjeet S. and Bickel, Peter J. and Yu, Bin},
  title   = {Metalearners for Estimating Heterogeneous Treatment Effects Using Machine Learning},
  journal = {Proceedings of the National Academy of Sciences},
  volume  = {116},
  number  = {10},
  pages   = {4156--4165},
  year    = {2019}
}

@article{nie2021quasi,
  author  = {Nie, Xinkun and Wager, Stefan},
  title   = {Quasi-Oracle Estimation of Heterogeneous Treatment Effects},
  journal = {Biometrika},
  volume  = {108},
  number  = {2},
  pages   = {299--319},
  year    = {2021}
}

@inproceedings{riquelme2018deep,
  author    = {Riquelme, Carlos and Tucker, George and Snoek, Jasper},
  title     = {Deep Bayesian Bandits Showdown: An Empirical Comparison of Bayesian Deep Networks for Thompson Sampling},
  booktitle = {International Conference on Learning Representations (ICLR)},
  publisher = {OpenReview.net},
  address   = {Vancouver, Canada},
  year      = {2018}
}

@inproceedings{wu2015algorithms,
  author    = {Wu, Huasen and Srikant, R. and Liu, Xin and Jiang, Chong},
  title     = {Algorithms with Logarithmic or Sublinear Regret for Constrained Contextual Bandits},
  booktitle = {Advances in Neural Information Processing Systems (NeurIPS)},
  pages     = {433--441},
  publisher = {Curran Associates Inc.},
  address   = {Red Hook, NY, USA},
  year      = {2015}
}

@inproceedings{li2010contextual,
  author    = {Li, Lihong and Chu, Wei and Langford, John and Schapire, Robert E.},
  title     = {A Contextual-Bandit Approach to Personalized News Article Recommendation},
  booktitle = {Proceedings of the 19th International Conference on World Wide Web (WWW)},
  pages     = {661--670},
  publisher = {Association for Computing Machinery},
  address   = {New York, NY, USA},
  year      = {2010}
}

@inproceedings{chu2011contextual,
  author    = {Chu, Wei and Li, Lihong and Reyzin, Lev and Schapire, Robert E.},
  title     = {Contextual Bandits with Linear Payoff Functions},
  booktitle = {International Conference on Artificial Intelligence and Statistics (AISTATS)},
  series    = {Proceedings of Machine Learning Research},
  volume    = {15},
  pages     = {208--214},
  publisher = {PMLR},
  address   = {Fort Lauderdale, FL, USA},
  year      = {2011}
}

@inproceedings{agrawal2013thompson,
  author    = {Agrawal, Shipra and Goyal, Navin},
  title     = {Thompson Sampling for Contextual Bandits with Linear Payoffs},
  booktitle = {International Conference on Machine Learning (ICML)},
  series    = {Proceedings of Machine Learning Research},
  volume    = {28},
  pages     = {127--135},
  publisher = {PMLR},
  address   = {Atlanta, GA, USA},
  year      = {2013}
}

@inproceedings{abbasi2011improved,
  author    = {Abbasi-Yadkori, Yasin and P{\'a}l, D{\'a}vid and Szepesv{\'a}ri, Csaba},
  title     = {Improved Algorithms for Linear Stochastic Bandits},
  booktitle = {Advances in Neural Information Processing Systems (NeurIPS)},
  pages     = {2312--2320},
  publisher = {Curran Associates Inc.},
  address   = {Red Hook, NY, USA},
  year      = {2011}
}

@inproceedings{dimakopoulou2019balanced,
  author    = {Dimakopoulou, Maria and Zhou, Zhengyuan and Athey, Susan and Imbens, Guido},
  title     = {Balanced Linear Contextual Bandits},
  booktitle = {Proceedings of the AAAI Conference on Artificial Intelligence},
  pages     = {3445--3453},
  publisher = {AAAI Press},
  address   = {Palo Alto, California, USA},
  year      = {2019}
}

@inproceedings{slivkins2023contextual,
  author    = {Slivkins, Aleksandrs and Sankararaman, Karthik Abinav and Foster, Dylan J.},
  title     = {Contextual Bandits with Packing and Covering Constraints: A Modular Lagrangian Approach via Regression},
  booktitle = {Conference on Learning Theory (COLT)},
  series    = {Proceedings of Machine Learning Research},
  volume    = {195},
  pages     = {4633--4656},
  publisher = {PMLR},
  address   = {Bangalore, India},
  year      = {2023}
}

@article{hadad2021confidence,
  author  = {Hadad, Vitor and Hirshberg, David A. and Zhan, Ruohan and Wager, Stefan and Athey, Susan},
  title   = {Confidence Intervals for Policy Evaluation in Adaptive Experiments},
  journal = {Proceedings of the National Academy of Sciences},
  volume  = {118},
  number  = {15},
  pages   = {e2014602118},
  year    = {2021}
}

@inproceedings{zhan2021offpolicy,
  author    = {Zhan, Ruohan and Hadad, Vitor and Hirshberg, David A. and Athey, Susan},
  title     = {Off-Policy Evaluation via Adaptive Weighting with Data from Contextual Bandits},
  booktitle = {Proceedings of the 27th ACM SIGKDD Conference on Knowledge Discovery and Data Mining (KDD)},
  pages     = {2125--2135},
  publisher = {Association for Computing Machinery},
  address   = {New York, NY, USA},
  year      = {2021}
}

@article{simchilevi2022bypassing,
  author  = {Simchi-Levi, David and Xu, Yunzong},
  title   = {Bypassing the Monster: A Faster and Simpler Optimal Algorithm for Contextual Bandits under Realizability},
  journal = {Mathematics of Operations Research},
  volume  = {47},
  number  = {3},
  pages   = {1904--1931},
  year    = {2022}
}

@inproceedings{chen2024contextual,
  author    = {Chen, Zhaohua and Ai, Rui and Yang, Mingwei and Pan, Yuqi and Wang, Chang and Deng, Xiaotie},
  title     = {Contextual Decision-Making with Knapsacks Beyond the Worst Case},
  booktitle = {Advances in Neural Information Processing Systems (NeurIPS)},
  publisher = {Curran Associates Inc.},
  pages = {88147--88193},
  address   = {Red Hook, NY, USA},
  year      = {2024}
}

@inproceedings{guo2025stochastic,
  author    = {Guo, Hengquan and Liu, Xin},
  title     = {On Stochastic Contextual Bandits with Knapsacks in Small Budget Regime},
  booktitle = {International Conference on Learning Representations (ICLR)},
  pages = {3647--3668},
  publisher = {OpenReview.net},
  address   = {Singapore},
  year      = {2025}
}

@inproceedings{lattimore2020learning,
  author    = {Lattimore, Tor and Szepesv{\'a}ri, Csaba and Weisz, Gell{\'e}rt},
  title     = {Learning with Good Feature Representations in Bandits and in {RL} with a Generative Model},
  booktitle = {International Conference on Machine Learning (ICML)},
  series    = {Proceedings of Machine Learning Research},
  volume    = {119},
  pages     = {5662--5670},
  publisher = {PMLR},
  address   = {Virtual Event},
  year      = {2020}
}

@inproceedings{zhou2020neural,
  author    = {Zhou, Dongruo and Li, Lihong and Gu, Quanquan},
  title     = {Neural Contextual Bandits with {UCB}-based Exploration},
  booktitle = {International Conference on Machine Learning (ICML)},
  series    = {Proceedings of Machine Learning Research},
  volume    = {119},
  pages     = {11492--11502},
  publisher = {PMLR},
  address   = {Virtual Event},
  year      = {2020}
}

@inproceedings{zhang2021neural,
  author    = {Zhang, Weitong and Zhou, Dongruo and Li, Lihong and Gu, Quanquan},
  title     = {Neural Thompson Sampling},
  booktitle = {International Conference on Learning Representations (ICLR)},
  publisher = {OpenReview.net},
  address   = {Virtual Event},
  year      = {2021}
}

@inproceedings{xu2022neural,
  author  = {Xu, Pan and Wen, Zheng and Zhao, Handong and Gu, Quanquan},
  title   = {Neural Contextual Bandits with Deep Representation and Shallow Exploration},
  booktitle = {International Conference on Learning Representations (ICLR)},
  publisher = {OpenReview.net},
  address = {Virtual Event},
  year    = {2022}
}

\clearpage
\appendix
\numberwithin{theorem}{section}
\numberwithin{assumption}{section}
\numberwithin{equation}{section}
\numberwithin{algorithm}{section}
\section*{Appendix}
The appendix contains the action-centered confidence and design results,
the budget-rate justification, the proof of Theorem~\ref{thm:variance},
the frozen neural variant, and the online neural-tangent extension with its
supporting proofs. The proof of Theorem~\ref{thm:regret} appears in
Section~\ref{sec:theory}.
\section{Action-Centered Least-Squares Confidence Set}
\label{app:posterior}

Let $D=Kd$, $\mathcal G_t=\sigma(\mathcal H_{t-1},X_t)$ be the pre-action
filtration, and $\mathcal F_t=\sigma(\mathcal G_t,A_t,Y_t)$ the post-outcome
filtration, with the offline inputs included in $\mathcal H_0$. Take $L>0$
and a predictable baseline estimate clipped to $[-F,F]$. Writing
$\zeta_t:=g_t-u_t^\top\theta^\star
=f(X_t)-\hat f(X_t)+\bar z_t^\top\theta^\star+\xi_t$, we have
$\|u_t\|_2\le2L$, $|\zeta_t|\le H:=2F+LS+\Xi$, and, conditional on
$\mathcal G_t$, $u_t$ is centered while
$\E[\xi_t\mid\mathcal G_t,A_t]=0$. Since the non-noise part of $\zeta_t$ is
$\mathcal G_t$-measurable, $u_t\zeta_t$ is $\mathcal F_t$-measurable with
conditional mean zero. Moreover, by Assumption~\ref{ass:subgauss} the noise $\xi_t$
is independent of $A_t$ given $\mathcal G_t$ and the remaining part of $\zeta_t$ is
$\mathcal G_t$-measurable, so $\zeta_t$ is conditionally independent of $u_t$ given
$\mathcal G_t$; this is the hypothesis under which the BOSE self-normalized inequality
(Lemmas 14, 16, 17 of \citet{krishnamurthy2018semiparametric}) controls
$\sum_t u_t\zeta_t$. The
normalizer nevertheless contains the same non-predictable $u_t$, so the ordinary
predictable-covariate theorem of \citet{abbasi2011improved} does not apply.

\begin{lemma}[Multi-action BOSE confidence set]
\label{lem:conf}
Set $q_T=4D\log(9T)+8\log(20T/\delta)$ and
$h_T=27D\log(1+2T/D)+54\log(20T/\delta)$. Under
Assumptions~\ref{ass:bounded}, \ref{ass:norms}, \ref{ass:realizability},
and~\ref{ass:subgauss}, if
$\alpha\ge4L^2q_T$, then, with probability at least $1-\delta/5$, simultaneously
for $t\le T$,
\[
\|\hat\theta_t-\theta^\star\|_{V_t}
\le H\sqrt{h_T}+\sqrt\alpha\,
\|\hat\theta^{(0)}-\theta^\star\|_2.
\]
\end{lemma}

Lemma~\ref{lem:conf} gives Eq.~\eqref{eq:radius} with any valid
$\bar\eta$. Since $\theta^\star$ lies in the radius-$S$ ball, projecting
$\hat\theta^{(0)}$ onto that ball gives the generic $\bar\eta=2S$; an external offline confidence
procedure may give a sharper data-dependent choice. Clipping $\hat f$ is used only to make
$|f-\hat f|\le2F$; action-centering, not clipping, removes its bias.

\begin{lemma}[Multi-action farthest-pair design]
\label{lem:farthest-design}
For any nonempty finite active set $\mathcal B$, features $z(a)$, and positive
definite $V$, the distribution in Algorithm~\ref{alg:sample-mix} satisfies, for
all $a,a'\in\mathcal B$,
\[
\|z(a)-z(a')\|_{V^{-1}}^2
\le4\,\E_{A\sim\pi}\|z(A)-\bar z\|_{V^{-1}}^2.
\]
\end{lemma}

\begin{corollary}[Mixture coverage]
\label{cor:mixture}
Fix $\gamma\in(0,1]$ and let
$\pi_t=(1-\gamma)q_t^{\mathrm{TS}}+\gamma\pi_t^{\mathrm{FP}}$, where $q_t^{\mathrm{TS}}$
is any predictable law on $\mathcal B_t$ and $\pi_t^{\mathrm{FP}}$ is the farthest-pair
law of Lemma~\ref{lem:farthest-design}. Then $\pi_t$ satisfies
Eq.~\eqref{eq:coverage} with $\kappa_0=\gamma/4$.
\end{corollary}
\begin{proof}
The round potential $w_t^\mu:=\E_{A\sim\mu}\|z_t(A)-\bar z_\mu\|_{V_{t-1}^{-1}}^2$ is
the $V_{t-1}^{-1}$-trace of $\mathrm{Cov}_\mu(z_t(A))$. For a mixture
$\mu=(1-\gamma)q+\gamma r$,
\[
\begin{aligned}
\mathrm{Cov}_\mu(z)
&=(1-\gamma)\mathrm{Cov}_q(z)+\gamma\,\mathrm{Cov}_r(z)\\
&\quad+\gamma(1-\gamma)(\mu_q-\mu_r)(\mu_q-\mu_r)^\top
\succeq\gamma\,\mathrm{Cov}_r(z),
\end{aligned}
\]
so $w_t^{\mathrm{mix}}\ge\gamma\,w_t^{\mathrm{FP}}$. By Lemma~\ref{lem:farthest-design},
every pair obeys $\|z_t(a)-z_t(a')\|_{V_{t-1}^{-1}}^2\le4w_t^{\mathrm{FP}}
\le(4/\gamma)\,w_t^{\mathrm{mix}}$, which is Eq.~\eqref{eq:coverage} with
$\kappa_0=\gamma/4$.
\end{proof}

\section{The Per-Period Budget Rate: Formal Justification}
\label{app:budget-rate}

The algorithm enforces the almost-sure cumulative cap
$\sum_t c(X_t,A_t)\le B$ through its affordable-action rule. The fluid benchmark
relaxes this cap to the expected-cost rate $\rho:=B/T$, making the Lagrangian
Eq.~\eqref{eq:lagrangian} time-homogeneous and its inner maximization
decompose pointwise across contexts as in Eq.~\eqref{eq:pi-lambda}. We show that
a stationary rate-feasible policy can be truncated to satisfy the pathwise cap,
the rate program has strong duality, and, for a fixed positive budget rate, the
fluid benchmark exceeds the optimal hard-budget value by at most
$\widetilde O(\sqrt T)$. Fix
$T$ and $B=T\rho$ with $\rho>0$; write
$\mathcal{F}_t := \sigma(X_1,A_1,\dots,X_t,A_t)$, $S_T := \sum_t c(X_t,A_t)$,
$r_{\max} := \sup_{x,a}|\tau_a(x)|<\infty$, and, by Assumption~\ref{ass:bounded},
$0\le c\le c_{\max}$. The rate program is
\begin{equation}\label{eq:p-rate}
  \text{(P-Rate)}\quad \max_\pi V(\pi)\ \text{s.t.}\ C(\pi)\le\rho,
\end{equation}
with $V,C$ as in Eq.~\eqref{eq:vc}. The cumulative program (P-Cum) maximizes
$\E[\sum_{t=1}^T\tau_{A_t}(X_t)]$ over nonanticipating policies subject to the
almost-sure cap $S_T\le B$.

\begin{lemma}[Concentration of cumulative spend]
\label{lem:concentration}
For any (possibly adaptive) policy and any $\delta\in(0,1)$, with probability at
least $1-\delta$,
$\max_{s\le T}\bigl|S_s-\sum_{t\le s}\E[c(X_t,A_t)\mid\mathcal F_{t-1}]\bigr|
\le c_{\max}\sqrt{2T\log(2/\delta)}$.
\end{lemma}

\begin{corollary}\label{cor:cum-bound}
If $\pi$ is stationary and rate-feasible, then with probability at least
$1-\delta$, simultaneously for $s\le T$,
$S_s\le s\rho+c_{\max}\sqrt{2T\log(2/\delta)}$.
\end{corollary}

\begin{proposition}[Truncation gives path-wise feasibility]
\label{prop:truncation}
Let $\pi$ be stationary with $C(\pi)\le\rho$. Draw fresh proposals
$A_t\sim\pi(\cdot\mid X_t)$ and let $\tilde\pi$ follow them until the first
proposal whose cost exceeds the remaining budget; from that round onward it
plays $a_0$. Then $\sum_t c(X_t,\tilde A_t)\le B$ almost surely and
$\big|T\,V(\pi)-\E[\sum_t\tau_{\tilde A_t}(X_t)]\big|
=O((r_{\max}c_{\max}/\rho)\sqrt{T\log(2T)})$.
\end{proposition}

\begin{proposition}[Strong duality of the rate-constrained program]
\label{prop:duality}
Under Assumptions~\ref{ass:iid} and~\ref{ass:bounded}, with bounded measurable
treatment effects as above, Slater's condition is met by
$\pi(a\mid x)=\one\{a=a_0\}$, since $c(\cdot,a_0)\equiv0$ and $\rho>0$. Program
Eq.~\eqref{eq:p-rate} is a convex program over stochastic kernels (equivalently,
occupancy measures $\nu(dx,a)=P_X(dx)\pi(a\mid x)$) and strong duality holds:
$\OPT_{\mathrm{LP}}(\rho)=\max_{C(\pi)\le\rho}V(\pi)
=\min_{\lambda\ge0}\max_\pi L(\pi,\lambda)$. At any optimal $\lambda^\star$ the
inner maximizer is pointwise,
$\pi^\star_{\lambda^\star}(x)\in\argmax_{a\in\A}(\tau_a(x)-\lambda^\star c(x,a))$.
A primal-optimal kernel is supported on these maximizers almost surely and may
need to randomize among tied actions to satisfy the cost constraint.
\end{proposition}

\begin{proposition}[Fluid LP versus hard-budget benchmark]
\label{prop:lp-gap}
Let $\OPT_{\mathrm{pathwise}}(B)$ be the optimum over all history-dependent
policies under the almost-sure cap $S_T\le B$. Under the assumptions of
Proposition~\ref{prop:duality},
$0\le T\cdot\OPT_{\mathrm{LP}}(\rho)-\OPT_{\mathrm{pathwise}}(B)
=O((r_{\max}c_{\max}/\rho)\sqrt{T\log(2T)})$, and hence the gap is $o(T)$
for fixed $r_{\max}$, $c_{\max}$, and $\rho>0$.
\end{proposition}

\section{Proofs of the Confidence and Design Lemmas}
\label{app:proof-lemmas}

\begin{proof}[Proof of Lemma~\ref{lem:conf}]
If $H=0$, then $\zeta_t=0$ almost surely and the stochastic term below vanishes.
Otherwise, the following argument bounds that term.
Normalize $Z_t:=u_t/(2L)$ and $\tilde\zeta_t:=\zeta_t/H$, so $\|Z_t\|_2\le1$ and
$|\tilde\zeta_t|\le1$. Conditional on $\mathcal G_t$ we have $\E[u_t\mid\mathcal G_t]=0$,
the non-noise part of $\zeta_t$ is $\mathcal G_t$-measurable, and
$\E[\xi_t\mid\mathcal G_t,A_t]=0$; hence $\E[u_t\zeta_t\mid\mathcal G_t]=0$ and
$\{u_t\zeta_t\}$ is a martingale-difference sequence adapted to $\{\mathcal F_t\}$.
Under a general action law, $Z_t$ need not be conditionally symmetric. Moreover,
it is realized after action sampling, so the predictable-covariate theorem of
\citet{abbasi2011improved} cannot be applied directly. Let
$\widehat\Sigma_t:=\sum_{s\le t}Z_sZ_s^\top$ and
$\Sigma_t:=\sum_{s\le t}\E[Z_sZ_s^\top\mid\mathcal G_s]$. Set
$a_T:=9D\log(9T)+8\log(20T/\delta)$ and
$\lambda_0:=\alpha/(4L^2)\ge q_T$. Lemma 14 of
\citet{krishnamurthy2018semiparametric}, applied at each prefix with failure
probability $\delta/(10T)$, gives
$\Sigma_t\preceq2\widehat\Sigma_t+a_TI$ simultaneously for $t\le T$
except on an event of probability at most $\delta/10$. Define
\[
 Q_T:=3\lambda_0-a_T
 \ge3D\log(9T)+16\log(20T/\delta)\ge1.
\]
On this covariance event,
\[
 \frac{V_t}{4L^2}=\lambda_0I+\widehat\Sigma_t
 \succeq\frac{Q_TI+\widehat\Sigma_t+\Sigma_t}{3}.
\]
Independently of conditioning on that event, their Lemma 16 applies to
$(Z_s,\tilde\zeta_s)$ with ridge $Q_TI$ and unit residual bound. A union bound
with failure probability $\delta/(10T)$ at each prefix, together with
$\mathrm{tr}(\widehat\Sigma_t+\Sigma_t)\le2T$ and $Q_T\ge1$, gives
\[
 \left\|\sum_{s\le t}Z_s\tilde\zeta_s\right\|_{
 (Q_TI+\widehat\Sigma_t+\Sigma_t)^{-1}}^2
 \le D\log(1+2T/D)+2\log(10T/\delta)
\]
simultaneously except on an event of probability at most $\delta/10$.
Intersecting the two events and rescaling therefore yields
\[
 \left\|\sum_{s\le t}u_s\zeta_s\right\|_{V_t^{-1}}^2
 \le H^2\{3D\log(1+2T/D)+6\log(10T/\delta)\}
 \le H^2h_T
\]
with total failure probability at most $\delta/5$. This is the empirical
normalizer argument underlying their Lemma 17. It holds for the adaptive,
$\mathcal G_t$-measurable affordable set, active set, and law $\pi_t$.
For the unprojected estimate
$\tilde\theta_t=V_t^{-1}(\alpha\hat\theta^{(0)}+\sum_{s\le t}u_sg_s)$ we have
$V_t(\tilde\theta_t-\theta^\star)=\alpha(\hat\theta^{(0)}-\theta^\star)+\sum_{s\le t}u_s\zeta_s$,
so by the triangle inequality and $V_t\succeq\alpha I$,
\[
\begin{aligned}
\|\tilde\theta_t-\theta^\star\|_{V_t}
&\le\Big\|\sum_{s\le t}u_s\zeta_s\Big\|_{V_t^{-1}}
+\alpha\|\hat\theta^{(0)}-\theta^\star\|_{V_t^{-1}}\\
&\le H\sqrt{h_T}+\sqrt\alpha\,\|\hat\theta^{(0)}-\theta^\star\|_2.
\end{aligned}
\]
The $V_t$-norm projection onto the radius-$S$ ball, which contains $\theta^\star$,
is non-expansive, so the same bound holds for the projected $\hat\theta_t$.
\end{proof}

\begin{proof}[Proof of Lemma~\ref{lem:farthest-design}]
The singleton case is immediate. Otherwise let $(a^+,a^-)$ attain the active-set
diameter $d^2:=\max_{a,a'\in\mathcal B}\|z(a)-z(a')\|_{V^{-1}}^2$, and let $\pi$
place mass $1/2$ on each of $a^+,a^-$. Then $\bar z=\tfrac12(z(a^+)+z(a^-))$, so
$z(a^\pm)-\bar z=\pm\tfrac12(z(a^+)-z(a^-))$ and each endpoint has squared centered
norm $d^2/4$; hence $\E_{A\sim\pi}\|z(A)-\bar z\|_{V^{-1}}^2=d^2/4$. Every pair
$a,a'\in\mathcal B$ obeys $\|z(a)-z(a')\|_{V^{-1}}^2\le d^2
=4\,\E_{A\sim\pi}\|z(A)-\bar z\|_{V^{-1}}^2$, which is Eq.~\eqref{eq:coverage} with
$\kappa_0=1/4$, independent of $|\mathcal B|$.
\end{proof}

\section{Proofs for the Per-Period Budget Rate}
\label{app:proof-budget}

\begin{proof}[Proof of Lemma~\ref{lem:concentration}]
The differences $c(X_t,A_t)-\E[c(X_t,A_t)\mid\mathcal F_{t-1}]$ form a
martingale-difference sequence bounded in absolute value by $c_{\max}$
(Assumption~\ref{ass:bounded}). Applying the maximal Azuma-Hoeffding inequality to
the partial sums $S_s-\sum_{t\le s}\E[c(X_t,A_t)\mid\mathcal F_{t-1}]$ gives, with
probability at least $1-\delta$, the uniform (in $s\le T$) deviation bound
$c_{\max}\sqrt{2T\log(2/\delta)}$.
\end{proof}

\begin{proof}[Proof of Corollary~\ref{cor:cum-bound}]
A stationary rate-feasible $\pi$ has $\E[c(X_t,A_t)\mid\mathcal F_{t-1}]=C(\pi)\le\rho$,
so $\sum_{t\le s}\E[c(X_t,A_t)\mid\mathcal F_{t-1}]\le s\rho$; substituting into
Lemma~\ref{lem:concentration} gives $S_s\le s\rho+c_{\max}\sqrt{2T\log(2/\delta)}$
simultaneously for $s\le T$.
\end{proof}

\begin{proof}[Proof of Proposition~\ref{prop:truncation}]
Couple $\tilde\pi$ to an untruncated sequence of proposals from $\pi$, and let
$S_s$ denote the spend of that proposal sequence. Take $\delta=(T+1)^{-2}$ in
Corollary~\ref{cor:cum-bound} and set
$D_T:=c_{\max}\sqrt{2T\log(2(T+1)^2)}$ and
$m:=\lceil(D_T+c_{\max})/\rho\rceil$.
On the resulting event, if $m<T$, every prefix with $s\le T-m$ satisfies
$S_s\le s\rho+D_T\le B-c_{\max}$. Thus no proposal is rejected before the
final $m$ rounds. If $m\ge T$, at most $T\le m$ rounds can change. On every
path, the actual actions follow affordable proposals until the first rejection
and then use the cost-free control, so cumulative spend never exceeds $B$.
Each changed round alters the incremental value by at most $r_{\max}$.
The expected reward of the untruncated sequence is $T V(\pi)$, so
\[
 \left|T V(\pi)-\E\!\left[\sum_t\tau_{\tilde A_t}(X_t)\right]\right|
 \le r_{\max}m+\frac{r_{\max}T}{(T+1)^2}
 =O\!\left(\frac{r_{\max}c_{\max}}{\rho}\sqrt{T\log(2T)}\right).
\]
\end{proof}

\begin{proof}[Proof of Proposition~\ref{prop:duality}]
Represent a kernel by its finitely many coordinates in
$L^\infty(P_X)^{K+1}$. The constraints $\pi_a\ge0$ and $\sum_a\pi_a=1$
define a weak-star closed subset of the product unit ball, which is compact by
Banach-Alaoglu. Since the costs and treatment effects are bounded and
measurable, $C$ and $V$ are weak-star continuous linear functionals.
Consequently, the achievable set of pairs $(C(\pi),V(\pi))$ is compact and
convex, and the rate-constrained maximum is attained. The strictly feasible
control-only kernel has $C=0<\rho$. A supporting-hyperplane argument for this
two-dimensional achievable set therefore gives strong duality and a finite
optimal multiplier:
\[
 \OPT_{\mathrm{LP}}(\rho)=\min_{\lambda\ge0}\max_\pi L(\pi,\lambda).
\]
For fixed $\lambda$, finite-action measurable maximization gives
\[
 \max_\pi L(\pi,\lambda)
 =\lambda\rho+\E_X\max_a\{\tau_a(X)-\lambda c(X,a)\}.
\]
Thus a deterministic pointwise maximizer attains the inner maximum.
At an optimal multiplier, a primal-optimal kernel assigns mass only to
pointwise maximizers and satisfies complementary slackness. Randomization
among ties can be necessary for primal feasibility; an arbitrary deterministic
tie choice need not satisfy the budget rate.
\end{proof}

\begin{proof}[Proof of Proposition~\ref{prop:lp-gap}]
For the upper bound, any history-dependent policy feasible for the almost-sure cap
$S_T\le B$ satisfies $\E[S_T]\le B$, so its time-averaged occupancy
$\bar\nu:=\tfrac1T\sum_t\nu_t$, where
$\nu_t(E,a):=\PP(X_t\in E,A_t=a)$, has context marginal $P_X$ and is
rate-feasible with
$V(\bar\nu)=\tfrac1T\E[\sum_t\tau_{A_t}(X_t)]$; hence
$\OPT_{\mathrm{pathwise}}(B)\le T\,\OPT_{\mathrm{LP}}(\rho)$. For the lower bound,
applying Proposition~\ref{prop:truncation} to a fluid-optimal stationary kernel
achieves expected cumulative value at least
$T\,\OPT_{\mathrm{LP}}(\rho)-O((r_{\max}c_{\max}/\rho)\sqrt{T\log(2T)})$
while satisfying the cap on every path. Combining the bounds proves the claim.
\end{proof}

\section{Proof of Theorem~\ref{thm:variance}}
\label{app:proof-variance}

\begin{proof}
Fix a round $t$ and condition on $(X_t,\mathcal H_{t-1})$; all expectations below
are over $A_t\sim\pi_t(\cdot\mid X_t)$ and the noise $\xi_t$. Recall
$Y_t=f(X_t)+\tau_{A_t}(X_t)+\xi_t$ with $\E[\xi_t\mid X_t,A_t]=0$ and
$\Var(\xi_t\mid X_t,A_t)=\sigma^2(X_t,A_t)$, the centered feature
$u_t=z_t(A_t)-\bar z_t$ with $\E[u_t\mid X_t,\mathcal H_{t-1}]=0$, and
$M_t=\E[u_tu_t^\top]$. Write $r_t=f(X_t)-\hat f(X_t)$, a constant given the
predictable $\hat f$, and $s_t^{\hat f}=u_t(Y_t-\hat f(X_t))$.

For part (a), since $Y_t-\hat f(X_t)=r_t+z_t(A_t)^\top\theta^\star+\xi_t$,
\[
\begin{aligned}
&\E[s_t^{\hat f}\mid X_t,\mathcal H_{t-1}]\\
&\quad= r_t\,\E[u_t] + \E[u_t z_t(A_t)^\top]\theta^\star + \E[u_t\xi_t] .
\end{aligned}
\]
The first term vanishes by $\E[u_t]=0$; the third vanishes because
$\E[\xi_t\mid X_t,A_t]=0$. For the second,
$\E[u_t z_t(A_t)^\top]=\E[u_t(u_t+\bar z_t)^\top]=\E[u_tu_t^\top]+\E[u_t]\bar z_t^\top=M_t$,
so $\E[s_t^{\hat f}\mid X_t,\mathcal H_{t-1}]=M_t\theta^\star$ for every predictable
$\hat f$, including $\hat f\equiv0$. The regularized least-squares estimator
Eq.~\eqref{eq:ls-update} solves the empirical version of
$M_t\theta=\E[s_t^{\hat f}]$. Lemma~\ref{lem:conf} gives
$|v_t^\top(\hat\theta_t-\theta^\star)|\le
\beta_t\|v_t\|_{V_t^{-1}}$. Thus every direction satisfying
$\beta_t\|v_t\|_{V_t^{-1}}\to0$ has vanishing estimation error on the confidence
event. Persistent exploration must make the width shrink to obtain this
conclusion. Directions used only by eliminated actions
need not remain excited, and full-parameter consistency is not claimed.

For part (b), with the i.i.d. logged sample of
Eq.~\eqref{eq:raw-comparator}, the population normal equations and
$\E[\xi\mid X,A]=0$ give
\[
\begin{aligned}
G_a b_a^{\mathrm{raw}}
&=\E[XY\mid A=a]\\
&=\E[Xf(X)\mid A=a]+G_a\theta_a^\star.
\end{aligned}
\]
Since $G_a\succ0$, it follows that
$b_a^{\mathrm{raw}}=\theta_a^\star+\gamma_a$ with
$\gamma_a=G_a^{-1}\E[Xf(X)\mid A=a]$. The control block has
$\theta_{a_0}^\star=0$, so the separate-arm fitted contrast at $x$ differs from
$\tau_a(x)$ by exactly $x^\top(\gamma_a-\gamma_{a_0})$. The law of large numbers
and continuity of matrix inversion give convergence of empirical arm-specific OLS
to these coefficients whenever every arm count diverges and the empirical second
moments converge to $G_a$. Its asymptotic bias under the common reference law
$P_X$ is therefore
$\|x^\top(\gamma_a-\gamma_{a_0})\|_{L^2(P_X)}$. If assignment is independent of
context, every arm has the same context law, hence the same $G_a$ and $\gamma_a$,
and the projection terms cancel. Under action-dependent sampling they can differ,
so a nonlinear $f$ outside the fitted linear class can produce a nonzero contrast
bias. This is a property of the specified separate-arm linear comparator, not of
the reward objective: Eq.~\eqref{eq:policy-equivalence} shows that a correctly
specified reward learner and the CATE learner share the same constrained optimum.

For part (c), write $m_t:=r_t+\tau_{A_t}(X_t)$, so
$s_t^{\hat f}=u_t(m_t+\xi_t)$ and, using $\E[\xi_t\mid X_t,A_t]=0$,
\[
\begin{aligned}
&\mathrm{Cov}(s_t^{\hat f}\mid X_t,\mathcal H_{t-1})\\
&\quad=\E[u_tu_t^\top\sigma^2(X_t,A_t)] + \mathrm{Cov}(u_t m_t) .
\end{aligned}
\]
The first term is the irreducible noise, independent of $\hat f$. For the second,
$u_t m_t=r_t u_t+u_t\tau_{A_t}$ with $r_t$ constant, so
\[
\begin{aligned}
\mathrm{Cov}(u_t m_t)
&= r_t^2\,\underbrace{\mathrm{Cov}(u_t)}_{=M_t}\\
&\quad+ 2 r_t\,\underbrace{\mathrm{Cov}(u_t,u_t\tau_{A_t})}_{=\E[u_tu_t^\top\tau_{A_t}]}\\
&\quad+ \mathrm{Cov}(u_t\tau_{A_t}),
\end{aligned}
\]
using $\E[u_t]=0$, which is Eq.~\eqref{eq:variance-decomp}; only the first two terms
depend on $\hat f$ through $r_t$. Taking traces and comparing the raw learner
($\hat f\equiv0$, $r_t=f(X_t)$) with the oracle ($\hat f=f$, $r_t=0$), the common
noise and $\mathrm{Cov}(u_t\tau_{A_t})$ terms cancel, leaving
\[
\begin{aligned}
&\mathrm{tr}\,\mathrm{Cov}(s_t^{0})-\mathrm{tr}\,\mathrm{Cov}(s_t^{f})\\
&\quad= f(X_t)^2\,\mathrm{tr}(M_t) + 2 f(X_t)\,\E[\|u_t\|_2^2\,\tau_{A_t}],
\end{aligned}
\]
which is Eq.~\eqref{eq:variance-trace}. This scalar is positive whenever
$|f(X_t)|\,\mathrm{tr}(M_t) > 2\,|\E[\|u_t\|_2^2\tau_{A_t}]|$; since
$|\tau_{A_t}|\le R_{\max}$ gives $|\E[\|u_t\|_2^2\tau_{A_t}]|\le R_{\max}\mathrm{tr}(M_t)$,
the condition $|f(X_t)|>2R_{\max}$ suffices when $\mathrm{tr}(M_t)>0$.
Thus residualizing the baseline strictly reduces the total score variance under
these two conditions. If $\mathrm{tr}(M_t)=0$, then $u_t=0$ almost surely
conditionally and both score covariances vanish. The matrix reduction is not one-directional in general
because of the cross term $2r_t\E[u_tu_t^\top\tau_{A_t}]$; the total-variance
statement is the correct one-directional guarantee.
\end{proof}

\section{Nonlinear Baselines and a Neural Variant}
\label{app:neural-ts}

Assumption~\ref{ass:realizability} restricts the CATEs to be linear in the raw
context. The baseline $f$ is already arbitrary: for any predictable $\hat f$,
action-centering removes $f-\hat f$ from the conditional moment, so a nonlinear or
misspecified baseline changes the confidence-radius constant but creates no
cumulative bias. We next analyze a frozen-feature counterpart of the neural
variant evaluated in Section~\ref{sec:exp-nonlinear}. Motivated by neural-linear
and deep-representation, shallow-exploration architectures
\citep{riquelme2018deep,xu2022neural}, it learns a neural representation from the
independent offline sample, freezes that representation before online
interaction, and updates only its linear head with action-centered feedback.
Conditional on the offline sample, freezing fixes both the feature map and the
head approximation used throughout the analysis. The resulting online features
are predictable; predictability alone would also permit features updated from
past observations.

\subsection{Frozen neural representation and algorithm}

Let $\widehat W$ be fitted only on the offline data and define an $m$-dimensional
one-hidden-layer representation
\[
  \phi(x):=m^{-1/2}\tanh(\widehat W x+\widehat b)\in\R^m.
\]
The normalization gives $\|\phi(x)\|_2\le1$; more generally, write
$\|\phi(x)\|_2\le L_\phi$. For the priced arms define
\[
  z_t^\phi(a):=e_a\otimes\phi(X_t)\in\R^{D_\phi},
  \qquad D_\phi:=Km,\qquad
  \bar z_t^\phi:=\sum_a\pi_t(a)z_t^\phi(a).
\]
The control block remains zero. We quantify the quality of the learned
representation by a uniform approximation error for a bounded linear head:
\begin{equation}
\begin{aligned}
  &\tau_a(x)=z^\phi(x,a)^\top\theta_\phi^\star+\varepsilon_a^\phi(x),
  \qquad\|\theta_\phi^\star\|_2\le S_\phi,\\
  &\sup_{x\in\operatorname{supp}(P_X),\,a\in\A}|\varepsilon_a^\phi(x)|
  \le\varepsilon_\phi,
\end{aligned}
  \label{eq:neural-approx}
\end{equation}
where $\varepsilon_{a_0}^\phi\equiv0$. Thus $\varepsilon_\phi=0$ is exact realizability in a
nonlinear, data-dependent feature space, not linearity in the raw context. The
analysis treats $\varepsilon_\phi$ as a certified input and does not infer an
approximation or finite-sample learning rate from network width or optimization.

\begin{algorithm}[t]
  \caption{Neural DR-CATE-CBwK with the analyzed farthest-pair rule.}
  \label{alg:neural-cbwk}
  \begin{algorithmic}[1]
    \Require Logged data $\D$, budget $B$, horizon $T$, hidden width $m$, and a
    valid representation-error bound $\varepsilon_\phi$.
    \State Cross-fit the nuisances and form the DR pseudo-outcomes
    $\psi_i(a)$ by Eq.~\eqref{eq:dr-pseudo}.
    \State Fit the multi-output network
    $\widehat\tau_a(x)=v_a^\top\phi_{\widehat W}(x)$ to offline contrast targets,
    such as $\psi_i(a)$; freeze $\phi:=\phi_{\widehat W}$, stack the fitted heads
    into $\widetilde\theta_\phi^{(0)}$, and fit the clipped baseline $\hat f$ as
    in Algorithm~\ref{alg:offline-dr}.
    \State Set
    $\hat\theta_\phi^{(0)}=
    \Proj_{\{\|\theta\|_2\le S_\phi\}}(\widetilde\theta_\phi^{(0)})$
    and clip the offline dual initialization to
    $\lambda_1\in[0,L_\phi S_\phi/c_{\min}]$.
    \State Set $V_0^\phi=\alpha I_{D_\phi}$,
    $b_0^\phi=\alpha\hat\theta_\phi^{(0)}$, $\hat\theta_{\phi,0}=\hat\theta_\phi^{(0)}$,
    and $S_0=0$.
    \For{$t=1,\ldots,T$}
      \State Observe $X_t$ and form the affordable set
      $\A_t^{\mathrm{aff}}$ as in Algorithm~\ref{alg:dr-cbwk}.
      \State Set
      $\hat s_t^\phi(a)=z_t^\phi(a)^\top\hat\theta_{\phi,t-1}
      -\lambda_t c(X_t,a)$ and form
      \begin{equation}
      \begin{aligned}
      \mathcal B_t^\phi=\Big\{a\in\A_t^{\mathrm{aff}}:\
      &\hat s_t^\phi(a')-\hat s_t^\phi(a)
      \le2\beta_t^\phi\|z_t^\phi(a')-z_t^\phi(a)\|_{(V_{t-1}^\phi)^{-1}}\\
      &+2\varepsilon_\phi,\quad\forall a'\in\A_t^{\mathrm{aff}}\Big\}.
      \end{aligned}
      \label{eq:neural-active}
      \end{equation}
      \State If $\mathcal B_t^\phi$ is a singleton, play its action. Otherwise,
      put probability $1/2$ on a farthest pair in $\mathcal B_t^\phi$ under
      $\|\cdot\|_{(V_{t-1}^\phi)^{-1}}$ and draw $A_t\sim\pi_t$.
      \State Observe $Y_t$, update spend, and set
      $u_t^\phi=z_t^\phi(A_t)-\bar z_t^\phi$ and
      $g_t=Y_t-\hat f(X_t)$.
      \State Update
      \[
      \begin{aligned}
      &V_t^\phi=V_{t-1}^\phi+u_t^\phi(u_t^\phi)^\top,\qquad
      b_t^\phi=b_{t-1}^\phi+u_t^\phi g_t,\\
      &\hat\theta_{\phi,t}=
      \argmin_{\|\theta\|_2\le S_\phi}
      \|\theta-(V_t^\phi)^{-1}b_t^\phi\|_{V_t^\phi}.
      \end{aligned}
      \]
      \State Set
      $\lambda_{t+1}=[\lambda_t+\eta_t(c(X_t,A_t)-\rho)]_+$.
    \EndFor
  \end{algorithmic}
\end{algorithm}

For $\varepsilon_\phi=0$, Eq.~\eqref{eq:neural-active} is exactly the active-set
rule of Algorithm~\ref{alg:sample-mix} in the learned feature space. For
$\varepsilon_\phi>0$, the additive slack prevents elimination caused solely by
representation error. As in the main paper, the farthest-pair rule is the
analyzed selector. The experiment uses a Thompson-style distribution supported
on the active set and calibrated confidence widths. The theorem below applies to
such a selector only when it uses the stated active set and confidence radius and
satisfies the coverage condition; the reported experiment is not presented as a
theorem-certified run.

\subsection{Confidence set and regret theorem}

The offline sample is independent of the online interaction. We therefore state the
result conditional on the fitted representation. This cleanly separates the online
regret argument from any procedure-specific guarantee for learning $\phi$.

\begin{assumption}[Frozen neural representation]\label{ass:neural-representation}
There is an event $\mathcal E_\phi$, measurable with respect to the independent
offline sample, on which $\phi$ is fixed before round one,
$0<L_\phi<\infty$, $\|\phi(X_t)\|_2\le L_\phi$ almost surely,
Eq.~\eqref{eq:neural-approx} holds, and the projected warm start satisfies
\[
  \|\hat\theta_\phi^{(0)}-\theta_\phi^\star\|_2\le\bar\eta_\phi.
\]
Here $\PP(\mathcal E_\phi)\ge1-\delta_{\mathrm{off}}$ and
$0\le\bar\eta_\phi\le2S_\phi$. The deterministic choice
$\bar\eta_\phi=2S_\phi$ is always valid on the representation event because both
$\theta_\phi^\star$ and $\hat\theta_\phi^{(0)}$ lie in the radius-$S_\phi$
ball. A sharper radius requires a representation-learner-specific offline
analysis.
\end{assumption}

\begin{assumption}[Neural active-set selection]\label{ass:neural-info}
Conditional on the frozen representation, the online selector draws a
$\mathcal G_t$-measurable distribution $\pi_t$ supported on
$\mathcal B_t^\phi$ such that, for a constant $\kappa_0>0$ and every
$a,a'\in\mathcal B_t^\phi$,
\begin{equation}
\|z_t^\phi(a)-z_t^\phi(a')\|_{(V_{t-1}^\phi)^{-1}}^2
\le \kappa_0^{-1}\,
\E\!\left[\|u_t^\phi\|_{(V_{t-1}^\phi)^{-1}}^2
\mid\mathcal G_t\right].
\label{eq:neural-coverage}
\end{equation}
The farthest-pair rule in Algorithm~\ref{alg:neural-cbwk} satisfies this
condition with $\kappa_0=1/4$ by Lemma~\ref{lem:farthest-design}.
\end{assumption}

Set
\[
\begin{aligned}
q_T^\phi&=4D_\phi\log(9T)+8\log(20T/\delta),\\
h_T^\phi&=27D_\phi\log(1+2T/D_\phi)+54\log(20T/\delta),
\end{aligned}
\]
$\alpha=4L_\phi^2q_T^\phi$, and
\begin{equation}
  \beta_t^\phi:=
  (\Xi+2F+L_\phi S_\phi)\sqrt{h_T^\phi}
  +\sqrt\alpha\,\bar\eta_\phi+\varepsilon_\phi\sqrt t.
  \label{eq:neural-radius}
\end{equation}

\begin{lemma}[Neural action-centered confidence set]
\label{lem:neural-confidence}
Under Assumptions~\ref{ass:bounded}, \ref{ass:subgauss}, and
\ref{ass:neural-representation}, conditional on $\mathcal E_\phi$, with
probability at least $1-\delta/5$ simultaneously for $0\le t\le T$,
\[
  \|\hat\theta_{\phi,t}-\theta_\phi^\star\|_{V_t^\phi}
  \le\beta_t^\phi.
\]
\end{lemma}

\begin{theorem}[Regret of the frozen neural variant]
\label{thm:neural-regret}
Suppose Assumptions~\ref{ass:iid}, \ref{ass:bounded},
\ref{ass:subgauss}, \ref{ass:neural-representation}, and
\ref{ass:neural-info} hold. Run Algorithm~\ref{alg:neural-cbwk} with
Eq.~\eqref{eq:neural-radius} and $\eta_t=\eta_0/\sqrt T$. Conditional on
$\mathcal E_\phi$, with probability at least $1-\delta$,
\begin{align}
R(T)\le\;&
\underbrace{C_1D_\phi\sqrt T\log\tfrac{KT}{\delta}}
_{R_{\mathrm{online}}}
+\underbrace{C_2\sqrt T}_{R_{\mathrm{dual}}}
+\underbrace{C_3D_\phi\sqrt T\,\bar\eta_\phi
\log\tfrac{KT}{\delta}}_{R_{\mathrm{warm}}}
\nonumber\\
&+\underbrace{C_4\left(\varepsilon_\phi T
\left[1+\sqrt{D_\phi\log\!\left(1+
\frac{4L_\phi^2T}{\alpha D_\phi}\right)}\right]
+\frac{\varepsilon_\phi^2T}{\sqrt\alpha}\right)}
_{R_{\mathrm{approx}}}.
\label{eq:neural-regret}
\end{align}
The constants depend on $\Xi$, $F$, $c_{\max}$, $c_{\min}$, $\rho$,
$L_\phi$, $S_\phi$, $\kappa_0$, $\eta_0$, and $\log(1/\delta)$, but not on
$T$, $m$, $K$, or $\varepsilon_\phi$. As in Theorem~\ref{thm:regret}, the
guarantee is for a fixed positive budget rate and is not uniform as
$B/T\to0$. Under exact neural realizability, $\varepsilon_\phi=0$, it gives
\[
  R(T)=\widetilde O\!\left(Km\sqrt T\right)
  +\widetilde O\!\left(Km\sqrt T\,\bar\eta_\phi\right).
\]
Thus the analyzed frozen-feature algorithm has the same horizon dependence as
the linear-CATE algorithm, with raw dimension $d$ replaced by hidden dimension
$m$, holding $L_\phi$ and $S_\phi$ fixed. Unconditionally, the bound holds with
probability at least $1-\delta-\delta_{\mathrm{off}}$.
\end{theorem}

The approximation term has two sources. The additive part is the direct
discrepancy between the true CATE and its linear head, while the square-root
factor arises when the misspecified head is estimated from action-centered
feedback. Two regimes should be distinguished. For fixed $D_\phi$ and bounded
$\bar\eta_\phi$, the condition $\varepsilon_\phi=o((D_\phi\log T)^{-1/2})$ is
sufficient only for sublinear regret, $R_{\mathrm{approx}}=o(T)$; it does not
preserve the statistical rate. To keep the approximation contribution no larger
than the semiparametric term $\widetilde O(D_\phi\sqrt T)$, and hence retain the
overall $\widetilde O(D_\phi\sqrt T)$ rate, one needs the stronger
$\varepsilon_\phi=\widetilde O(\sqrt{D_\phi/T})$. If $D_\phi$ grows with $T$,
one also needs $D_\phi(1+\bar\eta_\phi)\log(KT/\delta)=o(\sqrt T)$. A fixed
nonzero approximation error can produce linear regret without additional
structure, consistent with known sensitivity to feature misspecification
\citep{lattimore2020learning}. Assumption~\ref{ass:neural-representation}
therefore keeps the representation guarantee explicit rather than attributing a
finite-sample rate to the neural optimizer.

\subsection{Proof of Lemma~\ref{lem:neural-confidence}}

\begin{proof}
Condition throughout on $\mathcal E_\phi$, so $\phi$ and
$\hat\theta_\phi^{(0)}$ satisfy the required offline bounds before online
interaction. The claim at $t=0$ follows from
$V_0^\phi=\alpha I$ and the warm-start bound. For $t\ge1$, write
$\varepsilon_t^\phi=\varepsilon_{A_t}^\phi(X_t)$ and
\[
  \zeta_t^0:=
  f(X_t)-\hat f(X_t)
  +(\bar z_t^\phi)^\top\theta_\phi^\star+\xi_t.
\]
By Eq.~\eqref{eq:neural-approx},
\[
  g_t=(u_t^\phi)^\top\theta_\phi^\star+\zeta_t^0+\varepsilon_t^\phi.
\]
Conditional on $\mathcal G_t=\sigma(\mathcal H_{t-1},X_t)$, the non-noise part
of $\zeta_t^0$ is predictable and $\xi_t$ is independent of the sampled action.
Hence $\zeta_t^0$ is conditionally independent of $u_t^\phi$,
$\|u_t^\phi\|_2\le2L_\phi$, and
$|\zeta_t^0|\le\Xi+2F+L_\phi S_\phi$. The same BOSE
self-normalized inequality used in Lemma~\ref{lem:conf} gives, simultaneously
over $t$,
\begin{equation}
\left\|
\sum_{s\le t}u_s^\phi\zeta_s^0
\right\|_{(V_t^\phi)^{-1}}
\le
(\Xi+2F+L_\phi S_\phi)\sqrt{h_T^\phi}.
\label{eq:neural-bose-part}
\end{equation}
Also, $V_t^\phi\succeq\alpha I$ implies
\[
\left\|\alpha(\hat\theta_\phi^{(0)}-\theta_\phi^\star)
\right\|_{(V_t^\phi)^{-1}}
\le\sqrt\alpha\,\bar\eta_\phi.
\]

It remains to control the action-dependent approximation error, for which BOSE
conditional independence is unavailable. Let $U_t$ have rows
$(u_s^\phi)^\top$ and let
$\varepsilon_{1:t}^\phi=(\varepsilon_1^\phi,\ldots,\varepsilon_t^\phi)^\top$.
Since
$V_t^\phi=\alpha I+U_t^\top U_t$,
\[
\begin{aligned}
\left\|\sum_{s\le t}u_s^\phi\varepsilon_s^\phi\right\|_{(V_t^\phi)^{-1}}^2
&=(\varepsilon_{1:t}^\phi)^\top
U_t(V_t^\phi)^{-1}U_t^\top\varepsilon_{1:t}^\phi\\
&\le\|\varepsilon_{1:t}^\phi\|_2^2
\le t\varepsilon_\phi^2.
\end{aligned}
\]
If $\sigma_j$ are the singular values of $U_t$, the nonzero eigenvalues of
$U_t(\alpha I+U_t^\top U_t)^{-1}U_t^\top$ are
$\sigma_j^2/(\alpha+\sigma_j^2)\in[0,1)$, which proves the matrix inequality.

For the unprojected least-squares estimate $\widetilde\theta_{\phi,t}$, the
normal equations give
\[
V_t^\phi(\widetilde\theta_{\phi,t}-\theta_\phi^\star)
=\alpha(\hat\theta_\phi^{(0)}-\theta_\phi^\star)
+\sum_{s\le t}u_s^\phi\zeta_s^0
+\sum_{s\le t}u_s^\phi\varepsilon_s^\phi.
\]
The triangle inequality and the preceding three bounds yield
$\|\widetilde\theta_{\phi,t}-\theta_\phi^\star\|_{V_t^\phi}
\le\beta_t^\phi$. Projection onto the radius-$S_\phi$ ball in the
$V_t^\phi$ norm cannot increase the distance to
$\theta_\phi^\star$, which belongs to that ball. This proves the lemma.
\end{proof}

\subsection{Proof of Theorem~\ref{thm:neural-regret}}

\begin{proof}
All expectations and probabilities below are conditional on
$\mathcal E_\phi$. We follow the proof of Theorem~\ref{thm:regret} and make the
effects of representation error explicit. When $\varepsilon_\phi=0$, the
argument is, on $\mathcal E_\phi$, the linear-CATE proof with $(X_t,L,S,D)$
replaced by $(\phi(X_t),L_\phi,S_\phi,D_\phi)$. The additional steps below control the
action-dependent approximation residual, preserve the true optimum under
elimination, and track the resulting increase in the dual queue. Set
$G:=c_{\max}+\rho$ and
$R_{\max,\phi}:=L_\phi S_\phi+\varepsilon_\phi$, so
$|\tau_a(X_t)|\le R_{\max,\phi}$.

If $\varepsilon_\phi>1$, the deterministic bound
$R(T)\le2R_{\max,\phi}T\le2(L_\phi S_\phi+1)\varepsilon_\phi T$
proves Eq.~\eqref{eq:neural-regret} after increasing $C_4$. In the remainder,
assume $0\le\varepsilon_\phi\le1$.

We first show that the true affordable optimum remains active. Define the true
budget-adjusted score
\[
  s_t^\star(a):=\tau_a(X_t)-\lambda_t c(X_t,a)
  =z_t^\phi(a)^\top\theta_\phi^\star+\varepsilon_a^\phi(X_t)
  -\lambda_t c(X_t,a),
\]
and let $a_t^\star$ maximize it over $\A_t^{\mathrm{aff}}$. On the event of
Lemma~\ref{lem:neural-confidence}, the bound at time $t-1$ and monotonicity of
$\beta_t^\phi$ give, for every $a,a'$,
\[
\left|
(z_t^\phi(a)-z_t^\phi(a'))^\top
(\hat\theta_{\phi,t-1}-\theta_\phi^\star)
\right|
\le\beta_t^\phi
\|z_t^\phi(a)-z_t^\phi(a')\|_{(V_{t-1}^\phi)^{-1}}.
\]
For any $a'$ the optimality of $a_t^\star$ and
$|\varepsilon_{a'}^\phi(X_t)-\varepsilon_{a_t^\star}^\phi(X_t)|
\le2\varepsilon_\phi$ imply
\[
  \hat s_t^\phi(a')-\hat s_t^\phi(a_t^\star)
  \le\beta_t^\phi
  \|z_t^\phi(a')-z_t^\phi(a_t^\star)\|_{(V_{t-1}^\phi)^{-1}}
  +2\varepsilon_\phi.
\]
Therefore $a_t^\star\in\mathcal B_t^\phi$. Since the sampled action also lies in
$\mathcal B_t^\phi$, Eq.~\eqref{eq:neural-active} and one more use of the
confidence inequality yield
\begin{equation}
  s_t^\star(a_t^\star)-s_t^\star(A_t)
  \le3\beta_t^\phi
  \|z_t^\phi(a_t^\star)-z_t^\phi(A_t)\|_{(V_{t-1}^\phi)^{-1}}
  +4\varepsilon_\phi.
  \label{eq:neural-one-step}
\end{equation}

We next bound cumulative regret against the affordable optimum. By
Assumption~\ref{ass:neural-info},
\[
\|z_t^\phi(a_t^\star)-z_t^\phi(A_t)\|_{(V_{t-1}^\phi)^{-1}}^2
\le\kappa_0^{-1}w_t^\phi,
\quad
w_t^\phi:=
\E[\|u_t^\phi\|_{(V_{t-1}^\phi)^{-1}}^2\mid\mathcal G_t].
\]
To control the predictable potentials, set
$\Delta_t^\phi:=\|u_t^\phi\|_{(V_{t-1}^\phi)^{-1}}^2$ and
$b:=4L_\phi^2/\alpha\le1$. Then
$0\le\Delta_t^\phi\le b$ and
$\E[\Delta_t^\phi\mid\mathcal G_t]=w_t^\phi$.
Applying the exponential-supermartingale argument of
Eq.~\eqref{eq:potential-transfer} to the frozen neural features gives, with
probability at least $1-\delta/5$,
\[
\begin{aligned}
\sum_{t=1}^T w_t^\phi
&\le2\sum_{t=1}^T\Delta_t^\phi+2b\log(5/\delta)\\
&=O\!\left(
D_\phi\log\!\left(1+\frac{4L_\phi^2T}{\alpha D_\phi}\right)
+\log(1/\delta)\right).
\end{aligned}
\]
Here the realized sum is bounded by the elliptical potential inequality,
\[
\sum_{t=1}^T\Delta_t^\phi
\le2D_\phi\log\!\left(1+
\frac{4L_\phi^2T}{\alpha D_\phi}\right).
\]
Because $\beta_t^\phi$ is nondecreasing, summing
Eq.~\eqref{eq:neural-one-step} and applying Cauchy-Schwarz gives
\[
\begin{aligned}
R_{\mathrm{primal}}^{\mathrm{aff}}(T)
&\le
3\kappa_0^{-1/2}\beta_T^\phi\sqrt{T\sum_t w_t^\phi}
+4\varepsilon_\phi T\\
&=\widetilde O(D_\phi\sqrt T)
+\widetilde O(D_\phi\sqrt T\,\bar\eta_\phi)
+4\varepsilon_\phi T\\
&\quad+O\!\left(\varepsilon_\phi T
\sqrt{D_\phi\log\!\left(1+
\frac{4L_\phi^2T}{\alpha D_\phi}\right)}\right).
\end{aligned}
\]
This is where the approximation term enters the bandit rate. No term involving
$\|\hat f-f\|T$ appears: the shared baseline remains in $\zeta_t^0$, where
action-centering removes its conditional mean exactly.

We now control the dual variable, the prefix spend, and the affordable-set
restriction, the three places where the representation error $\varepsilon_\phi$
enters the budget analysis. Throughout, recall
$\alpha=4L_\phi^2q_T^\phi\ge4L_\phi^2>0$, $G=c_{\max}+\rho$, and
$R_{\max,\phi}=L_\phi S_\phi+\varepsilon_\phi$.

The dual term itself is unchanged. As in the derivation of
Eq.~\eqref{eq:regret-dual-main}, the update
$\lambda_{t+1}=[\lambda_t+\eta(c(X_t,A_t)-\rho)]_+$ with $\eta=\eta_0/\sqrt T$
satisfies
$\lambda_{t+1}^2\le\lambda_t^2+2\eta\lambda_t(c(X_t,A_t)-\rho)+\eta^2G^2$,
so telescoping gives
\[
R_{\mathrm{dual}}(T)=-\sum_t\lambda_t(c(X_t,A_t)-\rho)
\le\frac{\lambda_1^2}{2\eta}+\frac{\eta TG^2}{2}=O(\sqrt T)
\]
deterministically, independent of $\varepsilon_\phi$ and of the queue range.

The queue is uniformly bounded, with an $\varepsilon_\phi$-dependent range.
Projection onto the radius-$S_\phi$ ball and $V_{t-1}^\phi\succeq\alpha I$ give,
for every affordable $a$, $|z_t^\phi(a)^\top\hat\theta_{\phi,t-1}|\le L_\phi S_\phi$
and $\|z_t^\phi(a)\|_{(V_{t-1}^\phi)^{-1}}\le L_\phi/\sqrt\alpha$. Since the free
control has $z_t^\phi(a_0)=0$, $c(X_t,a_0)=0$, and
$\varepsilon_{a_0}^\phi\equiv0$, the rule
Eq.~\eqref{eq:neural-active} eliminates every priced action against $a_0$ once
\[
\lambda_t>\bar\lambda_\phi:=
\frac{L_\phi S_\phi+2\beta_T^\phi L_\phi/\sqrt\alpha+2\varepsilon_\phi}{c_{\min}}.
\]
On such a round $A_t=a_0$ and $\lambda_{t+1}=[\lambda_t-\eta\rho]_+\le\lambda_t$,
while any other round raises the queue by at most $\eta G$. With the
initialization $\lambda_1\le L_\phi S_\phi/c_{\min}$, induction gives
$0\le\lambda_t\le\Lambda_\phi$ for all $t$, where
\[
\Lambda_\phi:=\max\{\lambda_1,\ \bar\lambda_\phi+\eta G\}.
\]
Using $\bar\eta_\phi\le2S_\phi$ and $\varepsilon_\phi\le1$ in
Eq.~\eqref{eq:neural-radius},
$\beta_T^\phi/\sqrt\alpha=O(1)+O(\varepsilon_\phi\sqrt{T/\alpha})$, hence
\[
\Lambda_\phi=O(1)+O(\varepsilon_\phi\sqrt{T/\alpha}).
\]

The prefix spend obeys a deterministic bound. The nonnegative update gives
$\lambda_{t+1}/\eta\ge\lambda_t/\eta+(c(X_t,A_t)-\rho)$, so telescoping and
$0\le\lambda_{t+1}\le\Lambda_\phi$ yield
\[
\begin{aligned}
S_t=\sum_{s\le t}c(X_s,A_s)
&\le t\rho+\frac{\lambda_{t+1}-\lambda_1}{\eta}
\le t\rho+\frac{\Lambda_\phi}{\eta},\\
\frac{\Lambda_\phi}{\eta}
&=O(\sqrt T)+O\!\Big(\frac{\varepsilon_\phi T}{\sqrt\alpha}\Big).
\end{aligned}
\]
Thus $B-S_{t-1}>c_{\max}$, i.e. $\A_t^{\mathrm{aff}}=\A$, on all but the last
\[
m_\phi:=\Big\lceil\frac{\Lambda_\phi/\eta+c_{\max}}{\rho}\Big\rceil
=O(\sqrt T)+O\!\Big(\frac{\varepsilon_\phi T}{\sqrt\alpha}\Big)
\]
rounds.

The affordable optimum matches the fluid benchmark except on those rounds. The
master decomposition Eq.~\eqref{eq:regret-master-main} scores the global optimum over $\A$,
whereas Eq.~\eqref{eq:neural-one-step} bounds the gap against the affordable
optimum $a_t^\star$ over $\A_t^{\mathrm{aff}}$. On the $T-m_\phi$ rounds with
$\A_t^{\mathrm{aff}}=\A$ the two optima coincide; on each of the at most
$m_\phi$ remaining rounds the free control is affordable with adjusted score
zero, so replacing the global optimum by $a_t^\star$ costs at most
$R_{\max,\phi}$. The affordable-set restriction therefore adds at most
\[
R_{\max,\phi}\,m_\phi
=O(\sqrt T)+O(\varepsilon_\phi\sqrt T)
+O(\varepsilon_\phi T/\sqrt\alpha)+O(\varepsilon_\phi^2 T/\sqrt\alpha).
\]

The master martingale difference in Eq.~\eqref{eq:regret-master-main} is bounded by
$2(R_{\max,\phi}+\Lambda_\phi G)$, so Azuma-Hoeffding contributes
$O((R_{\max,\phi}+\Lambda_\phi G)\sqrt{T\log(1/\delta)})
=\widetilde O(\sqrt T)+O(\varepsilon_\phi T/\sqrt\alpha)$. Collecting the
preceding four displays, every $\varepsilon_\phi$-free term is
$\widetilde O(\sqrt T)$ and joins $C_2\sqrt T$. The $\varepsilon_\phi$-dependent
contributions are $\varepsilon_\phi\sqrt T$, $\varepsilon_\phi T/\sqrt\alpha$,
and $\varepsilon_\phi^2 T/\sqrt\alpha$; using $\sqrt T\le T$ and
$1/\sqrt\alpha\le1/(2L_\phi)$, the first two are
$O(\varepsilon_\phi T)$, with constants allowed to depend on $L_\phi$, and
together with the
$O(\varepsilon_\phi T\sqrt{D_\phi\log(1+4L_\phi^2T/(\alpha D_\phi))})$ term
from $R_{\mathrm{primal}}^{\mathrm{aff}}$ and the residual
$\varepsilon_\phi^2 T/\sqrt\alpha$ they form $R_{\mathrm{approx}}$ in
Eq.~\eqref{eq:neural-regret}. The initial case distinction covers all
$\varepsilon_\phi\ge0$. The
free control is affordable on every round, so $\sum_t c(X_t,A_t)\le B$ holds
pathwise on every sample path.

Inserting these bounds into the weak-duality master decomposition
Eq.~\eqref{eq:regret-master-main} completes the argument. Conditional on
$\mathcal E_\phi$, the confidence event, master martingale event, and
predictable-to-realized potential event each fail with probability at most
$\delta/5$; the dual and pathwise-feasibility bounds are deterministic. A union
bound proves Eq.~\eqref{eq:neural-regret} with probability at least $1-\delta$.
Finally, $\PP(\mathcal E_\phi^c)\le\delta_{\mathrm{off}}$ gives the
unconditional probability. Setting $\varepsilon_\phi=0$ removes every
approximation term and proves the exact-realizability statement.
\end{proof}

The theorem applies to any predictable representation fixed before online
interaction. Assumption~\ref{ass:neural-representation} supplies
$\varepsilon_\phi$ and $\bar\eta_\phi$ as certified offline inputs; the result
does not derive them from a particular network width, sample size, or optimizer.
Unlike NeuralUCB and Neural Thompson Sampling
\citep{zhou2020neural,zhang2021neural}, it also does not update the representation
during interaction.

The next section analyzes a complementary online construction: the nonlinear
network is retrained after each round, while initialization tangent features
provide its confidence geometry.

\section{An Online Action-Centered Neural-Tangent Extension}
\label{app:neural-ntk}

This section develops a separate neural-tangent extension in which the network is
retrained after every observation, while exploration uses gradients frozen at
initialization. The construction differs from the frozen representation of
Appendix~\ref{app:neural-ts}: the complete nonlinear network supplies the online
score, while its confidence geometry is measured by the information gain of an
action-centered neural tangent kernel (NTK). The key device is balanced
farthest-pair sampling. It converts action-centering into a Rademacher pseudo-outcome
with a predictable tangent contrast, which permits a kernel self-normalized
argument without applying an ordinary linear-bandit theorem to a feature observed
only after action sampling.

The result gives an online neural-network algorithm with a conditional NTK regret
guarantee for nonlinear CATE functions in the initialization tangent-feature class. It
combines causal action-centering, nonlinear network scoring, fixed centered-kernel
information gain, and a hard budget constraint. The regret theorem isolates one
direct centered-loss coupling event between nonlinear pair predictions and the
fixed-NTK ridge predictor. This avoids requiring a network to track a ridge
estimator assembled from different historical parameter iterates. It is not an
end-to-end finite-width theorem until that event is verified from explicit
architecture, width, learning-rate, and optimization assumptions.

\subsection{Inherited setting and NTK-specific scope}

We retain the action set, outcome decomposition, cost conditions, budget rate,
policy value $V(\pi)$, policy cost $C(\pi)$, fluid benchmark
$\OPT_{\mathrm{LP}}(\rho)$, and regret $R(T)$ from
Section~\ref{sec:setup}. Thus Assumptions~\ref{ass:iid}, \ref{ass:bounded}, and
\ref{ass:subgauss} continue to hold. The prefit baseline $\hat f$ is constructed
as in Section~\ref{sec:algorithm}, independently of the online sample and clipped
to $[-F,F]$. Only linear CATE realizability is replaced by the nonlinear
condition below.

Let $\mathcal H_0^{\mathrm N}$ contain the initialization $\theta_0$, the prefit
baseline $\hat f$, and any other pre-online algorithmic randomness, all
independent of the online sequence, and define
\[
 \mathcal H_t^{\mathrm N}
 :=\sigma(\mathcal H_0^{\mathrm N},X_1,A_1,Y_1,\ldots,X_t,A_t,Y_t).
\]
The conditional statements in Assumption~\ref{ass:subgauss} are understood with
this augmented history. At the end of round $t$, the algorithmic state
$(\theta_t,V_t^0,b_t^0,\widehat w_t^0,\lambda_{t+1},S_t)$ is
$\mathcal H_t^{\mathrm N}$-measurable.

\begin{assumption}[Bounded nonlinear CATE]
\label{ass:ntk-cate}
The treatment contrasts satisfy
\[
 \sup_{x\in\operatorname{supp}(P_X),a\in\A}|\tau_a(x)|
 \le R_{\mathrm{NTK}}.
\]
\end{assumption}

The extension differs from the frozen neural variant because the complete network
is retrained online, although confidence widths retain the initialization tangent
geometry. It also differs from Neural Thompson Sampling. Neural Thompson
Sampling regresses observed rewards on single-action network outputs and samples
Gaussian reward scores. The method below regresses action-centered pairwise
CATE observations on tangent contrasts and uses a balanced farthest-pair rule.
The symmetry of that rule, rather than posterior sampling, removes the shared
baseline exactly.

\subsection{Online network and balanced action-centering}

Let $h:\A\times\X\times\R^p\to\R$ be an action-conditioned
$L_{\mathrm{net}}$-layer fully connected network of width $m$ with shared
parameters $\theta\in\R^p$. The
action can be encoded by a one-hot input or an equivalent multi-output head. We
hard-code $h(a_0,x;\theta)=0$. The initialization $\theta_0$ is an algorithm
input independent of the data and satisfies $h(a,x;\theta_0)=0$. For example,
this identity follows by duplicating each initialized hidden feature with
opposite output weights. The theorem requires no further initialization law
beyond Assumption~\ref{ass:ntk-stability}. Define the initialization tangent
feature
\begin{equation}
 g_0(a,x):=m^{-1/2}\nabla_\theta h(a,x;\theta_0),
 \qquad g_0(a_0,x):=0.
 \label{eq:ntk-initial-feature}
\end{equation}
At the start of round $t$, the current parameters $\theta_{t-1}$ are
$\mathcal H_{t-1}^{\mathrm N}$-measurable, but the confidence feature is
$g_0(a):=g_0(a,X_t)$. Thus $g_0(a)$ is predictable after observing $X_t$ and
before drawing $A_t$. The network value $h(a,X_t;\theta_{t-1})$, rather than a
frozen linear head, is the online CATE estimate.

Suppose a predictable active set $\mathcal B_t$ contains at least two actions.
Choose a pair $(a_t^+,a_t^-)$ from this set, draw an independent Rademacher sign
$\sigma_t\in\{-1,+1\}$, and play $a_t^+$ when $\sigma_t=+1$ and $a_t^-$ when
$\sigma_t=-1$. Define
\begin{equation}
\begin{aligned}
 d_t^0&:=g_0(a_t^+)-g_0(a_t^-),\\
 \widetilde Y_t&:=2\sigma_t\{Y_t-\hat f(X_t)\}.
\end{aligned}
\label{eq:ntk-pair-observation}
\end{equation}
If the active set is the singleton $\{a\}$, set $a_t^+=a_t^-=a$, play $a$, and
set $d_t^0=0$ and $\widetilde Y_t=0$. With
$V_0^0=\alpha_{\mathrm{NTK}}I$ and $b_0^0=0$,
maintain the initialization-tangent ridge quantities
\begin{equation}
 V_t^0=V_{t-1}^0+d_t^0(d_t^0)^\top,
 \qquad b_t^0=b_{t-1}^0+d_t^0\widetilde Y_t,
 \qquad \widehat w_t^0=(V_t^0)^{-1}b_t^0.
 \label{eq:ntk-ridge-update}
\end{equation}
These quantities are used in the proof and in the confidence radius. The action
score itself remains the nonlinear network output.

After round $t$, retrain the full network from $\theta_0$ for $J$ gradient steps
on the regularized pairwise loss
\begin{equation}
 \mathcal L_t(\theta)
 :=\frac12\sum_{s\le t}
 \left(h(a_s^+,X_s;\theta)-h(a_s^-,X_s;\theta)
 -\widetilde Y_s\right)^2
 +\frac{m\alpha_{\mathrm{NTK}}}{2}\|\theta-\theta_0\|_2^2.
 \label{eq:ntk-centered-loss}
\end{equation}
The scaling connecting this objective to Eq.~\eqref{eq:ntk-ridge-update} is
\begin{equation}
 w(\theta):=\sqrt m\,(\theta-\theta_0).
 \label{eq:ntk-parameter-scaling}
\end{equation}
Because the initialization output is zero, first-order expansion of the pair
output gives
\[
 h(a_s^+,X_s;\theta)-h(a_s^-,X_s;\theta)
 =(d_s^0)^\top w(\theta)+\operatorname{Rem}_s(\theta).
\]
Moreover,
\[
 \frac{m\alpha_{\mathrm{NTK}}}{2}\|\theta-\theta_0\|_2^2
 =\frac{\alpha_{\mathrm{NTK}}}{2}\|w(\theta)\|_2^2.
\]
Discarding the Taylor remainders therefore gives the linearized objective
\begin{equation}
 \mathcal L_t^0(w)
 :=\frac12\sum_{s\le t}
 \left((d_s^0)^\top w-\widetilde Y_s\right)^2
 +\frac{\alpha_{\mathrm{NTK}}}{2}\|w\|_2^2,
 \qquad
 \argmin_w\mathcal L_t^0(w)=\widehat w_t^0.
 \label{eq:ntk-linearized-loss}
\end{equation}
The last identity follows from the normal equations and shows that the ridge
parameter is the exact optimizer of the initialization-linearized neural loss,
not an unrelated auxiliary estimator.
The next round therefore uses a newly trained nonlinear score, while retaining the
initialization features for uncertainty. The guarantee remains an NTK guarantee
because its realizability and confidence conditions are stated in tangent-feature
space.
Restarting from $\theta_0$ and fitting all observations at every round is a
theoretical construction chosen to simplify comparison with the initialization
NTK. It uses $J\sum_{t=1}^T t=J T(T+1)/2$ example-gradient evaluations, apart
from network-evaluation costs. A warm-started implementation would be cheaper,
but requires a different stability lemma and is not analyzed here.

Given a symmetric pairwise confidence radius $r_t(a,a')$, later instantiated in
Eq.~\eqref{eq:ntk-pair-radius}, define the residual spend and affordable set by
$S_{t-1}=\sum_{s<t}c(X_s,A_s)$ and
$\A_t^{\mathrm{aff}}=\{a:c(X_t,a)\le B-S_{t-1}\}$. The budget-adjusted network
score and active set are defined below. If the active set is empty, the algorithm
plays the control. When the active set has at least two actions, its farthest
distinct pair is also defined below. All maximization ties are resolved by a
fixed deterministic ordering. Hence the active set and selected pair are
measurable with respect to $\sigma(\mathcal H_{t-1}^{\mathrm N},X_t)$.
\begin{equation}
\begin{aligned}
 \widehat s_t(a)
 &:=h(a,X_t;\theta_{t-1})-\lambda_t c(X_t,a),\\
 \mathcal B_t^{\mathrm{NTK}}
 &:=\Big\{a\in\A_t^{\mathrm{aff}}:\
 \widehat s_t(a')-\widehat s_t(a)\le2r_t(a,a'),
 \forall a'\in\A_t^{\mathrm{aff}}\Big\},\\
 (a_t^+,a_t^-)
 &\in\argmax_{\substack{a,a'\in\mathcal B_t^{\mathrm{NTK}}\\a\ne a'}}
 \|g_0(a)-g_0(a')\|_{(V_{t-1}^0)^{-1}}.
\end{aligned}
\label{eq:ntk-active-pair}
\end{equation}

\begin{algorithm}[t]
 \caption{Action-Centered NTK-CBwK with balanced pairwise exploration.}
 \label{alg:ntk-cbwk}
 \begin{algorithmic}[1]
   \Require Budget $B$, horizon $T$, baseline estimate $\hat f$, initialized
  network $h(\cdot,\cdot;\theta_0)$, ridge $\alpha_{\mathrm{NTK}}$, and dual
  step $\eta_0$.
   \Require Neural step size $\gamma$, gradient steps $J$, initial dual
   $\lambda_1\ge0$, and radius $r_t$ from Eq.~\eqref{eq:ntk-pair-radius}.
  \State Set $\rho\gets B/T$, $V_0^0\gets\alpha_{\mathrm{NTK}}I_p$,
   $b_0^0\gets0$,
   $\widehat w_0^0\gets0$, and $S_0\gets0$.
  \For{$t=1,\ldots,T$}
   \State Observe $X_t$ and compute $g_0(a)=g_0(a,X_t)$ from
   Eq.~\eqref{eq:ntk-initial-feature} for every $a\in\A$.
   \State Set
   $\A_t^{\mathrm{aff}}\gets
   \{a\in\A:c(X_t,a)\le B-S_{t-1}\}$.
  \State Compute $\widehat s_t$ and $\mathcal B_t^{\mathrm{NTK}}$ using
  Eq.~\eqref{eq:ntk-active-pair}.
  \If{$\mathcal B_t^{\mathrm{NTK}}=\varnothing$}
   \State Set $a_t^+=a_t^-=A_t\gets a_0$, $d_t^0\gets0$, and
   $\widetilde Y_t\gets0$.
  \ElsIf{$\mathcal B_t^{\mathrm{NTK}}=\{a\}$}
    \State Set $a_t^+=a_t^-=A_t\gets a$, $d_t^0\gets0$, and
    $\widetilde Y_t\gets0$.
   \Else
   \State Choose $(a_t^+,a_t^-)$ by the farthest-pair rule in
   Eq.~\eqref{eq:ntk-active-pair}.
    \State Draw $\sigma_t$ uniformly from $\{-1,+1\}$ and set
    $A_t\gets a_t^+$ if $\sigma_t=+1$, and $A_t\gets a_t^-$ otherwise.
   \EndIf
   \State Observe $Y_t$ and set
   $S_t\gets S_{t-1}+c(X_t,A_t)$.
   \If{$a_t^+\ne a_t^-$}
    \State Set $d_t^0\gets g_0(a_t^+)-g_0(a_t^-)$ and
    $\widetilde Y_t\gets2\sigma_t\{Y_t-\hat f(X_t)\}$.
   \EndIf
     \State Update $(V_t^0,b_t^0,\widehat w_t^0)$ by
   Eq.~\eqref{eq:ntk-ridge-update}.
   \State Set $\theta_t^{(0)}\gets\theta_0$.
   \For{$j=0,\ldots,J-1$}
    \State $\theta_t^{(j+1)}\gets\theta_t^{(j)}
    -\gamma\nabla_\theta\mathcal L_t(\theta_t^{(j)})$.
   \EndFor
   \State Set $\theta_t\gets\theta_t^{(J)}$.
   \State Set
   $\lambda_{t+1}\gets
   [\lambda_t+(\eta_0/\sqrt T)\{c(X_t,A_t)-\rho\}]_+$.
  \EndFor
 \end{algorithmic}
\end{algorithm}

The algorithm maintains two coupled objects. The network output supplies the
nonlinear action score, while $(V_t^0,b_t^0,\widehat w_t^0)$ supplies fixed
initialization confidence geometry used to certify network contrasts. The next
propositions establish
that balanced pairwise feedback targets the CATE difference and preserves the
positive-semidefinite NTK structure.

\begin{proposition}[Exact action-centered pseudo-outcome]
\label{prop:ntk-centered-identity}
Let $\mathcal G_t=\sigma(\mathcal H_{t-1}^{\mathrm N},X_t,a_t^+,a_t^-)$, and
suppose Assumptions~\ref{ass:bounded}, \ref{ass:subgauss}, and
\ref{ass:ntk-cate} hold. Then
\begin{equation}
 \E[\widetilde Y_t\mid\mathcal G_t]
 =\tau_{a_t^+}(X_t)-\tau_{a_t^-}(X_t).
 \label{eq:ntk-centered-identity}
\end{equation}
Moreover,
\begin{equation}
 \widetilde Y_t
 =\tau_{a_t^+}(X_t)-\tau_{a_t^-}(X_t)+\nu_t,
 \qquad
 \E[\nu_t\mid\mathcal G_t]=0,
 \label{eq:ntk-pseudo-noise}
\end{equation}
where $|\nu_t|\le R_{\mathrm{ac}}:=2(2F+R_{\mathrm{NTK}}+\Xi)$ almost surely.
\end{proposition}

\begin{proof}
The singleton case has $a_t^+=a_t^-$ and both sides of
Eq.~\eqref{eq:ntk-centered-identity} are zero. Otherwise, write
$r_t^f=f(X_t)-\hat f(X_t)$ and
$\tau_t^\pm=\tau_{a_t^\pm}(X_t)$. Balanced sampling gives
\[
 \tau_{A_t}(X_t)
 =\frac{\tau_t^++\tau_t^-}{2}
 +\frac{\sigma_t}{2}(\tau_t^+-\tau_t^-).
\]
Substitution into $2\sigma_t(Y_t-\hat f(X_t))$ yields
\[
 \widetilde Y_t
 =\tau_t^+-\tau_t^-
 +2\sigma_t\left(r_t^f+\frac{\tau_t^++\tau_t^-}{2}+\xi_t\right).
\]
Given $\mathcal G_t$ and a distinct pair, $A_t$ is a one-to-one function of
$\sigma_t$. Conditional independence of $\xi_t$ from the sampled action
therefore gives
$\E[\sigma_t\xi_t\mid\mathcal G_t]=0$. The remaining terms in parentheses are
$\mathcal G_t$-measurable and $\E[\sigma_t\mid\mathcal G_t]=0$. Hence the last
term is a martingale difference. Clipping gives $|r_t^f|\le2F$, while
$|\tau_t^\pm|\le R_{\mathrm{NTK}}$ and $|\xi_t|\le\Xi$, which proves the bound.
\end{proof}

Proposition~\ref{prop:ntk-centered-identity} is the main compatibility result
between action-centering and NTK exploration. The random sign moves into the
response, while the initialization contrast $d_t^0$ is fixed before sampling.
Consequently, standard predictable-covariate kernel concentration applies to the
sequence $(d_t^0,\widetilde Y_t)$. A general nonbalanced action law instead
produces a post-action centered feature and would require a kernel-valued extension
of the BOSE self-normalized inequality.

\begin{proposition}[Centering preserves the NTK structure]
\label{prop:ntk-centering-kernel}
Let $K_0((x,a),(x',a'))=\langle g_0(a,x),g_0(a',x')\rangle$ be the
initialization NTK. For pair contrasts $q=(x,a^+,a^-)$ define
\begin{equation}
 K_0^{\mathrm{ac}}(q,q')
 :=\left\langle
 g_0(a^+,x)-g_0(a^-,x),
 g_0(a'^+,x')-g_0(a'^-,x')
 \right\rangle.
 \label{eq:centered-ntk}
\end{equation}
Then $K_0^{\mathrm{ac}}$ is positive semidefinite. For any finite collection of
pairs, its Gram matrix is $C H_0 C^\top$, where $H_0$ is the uncentered NTK Gram
matrix over the participating action-context inputs and each row of $C$ is the
corresponding signed contrast.
\end{proposition}

\begin{proof}
For any coefficients $v_1,\ldots,v_n$,
\[
 \sum_{i,j}v_iv_jK_0^{\mathrm{ac}}(q_i,q_j)
 =\left\|\sum_i v_i
 \{g_0(a_i^+,x_i)-g_0(a_i^-,x_i)\}\right\|_2^2\ge0.
\]
The matrix identity follows by expanding the two signed differences. Thus
centering changes the kernel by a linear contrast operator and does not destroy
positive semidefiniteness.
\end{proof}

\subsection{NTK assumptions and confidence set}

The following assumption states initialization-feature realizability and isolates
the architecture-specific step that transfers fixed-NTK ridge confidence to the
trained nonlinear pair predictor. Under the usual limiting-kernel interpretation,
the first condition is the corresponding centered NTK RKHS condition. Unlike a
moving-gradient formulation, no comparison among historical tangent geometries is
needed.

For a deterministic radius $R_\theta$, define the NTK neighborhood
\[
 \mathcal N_\theta
 :=\{\theta:\|w(\theta)\|_2\le R_\theta\}.
\]
For each $t\ge1$, let $\bar\theta_t$ be a selected element of
$\argmin_{\theta\in\mathcal N_\theta}\mathcal L_t(\theta)$, and set
$\bar\theta_0=\theta_0$. Write $w_t=w(\theta_t)$ and
$\bar w_t=w(\bar\theta_t)$ using Eq.~\eqref{eq:ntk-parameter-scaling}. This
reference definition does not assume convergence of the gradient-descent
parameter sequence.

\begin{assumption}[Modular fixed-NTK verification conditions]
\label{ass:ntk-stability}
The bounds and error tolerances below are deterministic and nonnegative, with
$L_g>0$ and $\alpha_{\mathrm{NTK}}>0$.
For $P_{\theta_0}$-almost every initialization, there is a vector $w^\star$,
measurable with respect to $\theta_0$ and therefore selected before online
interaction, such that $\|w^\star\|_2\le S_{\mathrm{NTK}}$ and, for a
deterministic $\epsilon_{\mathrm{init}}(m)$,
\[
 \left|\tau_a(x)-g_0(a,x)^\top w^\star\right|
 \le\epsilon_{\mathrm{init}}(m)
\]
for $P_X$-almost every $x$ and every $a\in\A$.

Under the joint law of the initialization and online trajectory, there are events
$\mathcal E_{\mathrm{env}}$, $\mathcal E_{\mathrm{lin}}$,
$\mathcal E_{\mathrm{id}}$, and $\mathcal E_{\mathrm{opt}}$ whose failure
probabilities are at most $\delta_{\mathrm{env}}$,
$\delta_{\mathrm{lin}}$, $\delta_{\mathrm{id}}$, and
$\delta_{\mathrm{opt}}$, respectively, with the following properties.

\begin{enumerate}
 \item On $\mathcal E_{\mathrm{env}}$, the selected minimizers
 $\bar\theta_t$ exist, every gradient-descent iterate remains in
 $\mathcal N_\theta$, and initialization gradients and trained outputs satisfy
 \[
 \begin{aligned}
 \sup_{a,x}\|g_0(a,x)\|_2&\le L_g,\\
 \sup_{t,a,x}|h(a,x;\theta_{t-1})|&\le R_h,
 \end{aligned}
 \]
 where the suprema use $t\le T$, $a\in\A$, and
 $x\in\operatorname{supp}(P_X)$.

 \item On $\mathcal E_{\mathrm{lin}}$, uniformly over $t\le T$,
 $a,a'\in\A$, and $x\in\operatorname{supp}(P_X)$,
 \[
 \left|h(a,x;\theta_{t-1})-h(a',x;\theta_{t-1})
 -\{g_0(a,x)-g_0(a',x)\}^\top w_{t-1}\right|
 \le\epsilon_{\mathrm{lin}}(m,T).
 \]

 \item On $\mathcal E_{\mathrm{opt}}$, uniformly over the same indices,
 \[
 \left|\{g_0(a,x)-g_0(a',x)\}^\top
 (w_{t-1}-\bar w_{t-1})\right|
 \le\epsilon_{\mathrm{opt}}(m,T,J).
 \]

 \item On $\mathcal E_{\mathrm{id}}$, uniformly over the same indices,
 \[
 \left|\{g_0(a,x)-g_0(a',x)\}^\top
 (\bar w_{t-1}-\widehat w_{t-1}^0)\right|
 \le\epsilon_{\mathrm{id}}(m,T).
 \]

 \item There is a deterministic capacity $\overline\Gamma_T^0$ such that,
 for every initialization in a $P_{\theta_0}$-probability-one set and uniformly
 over all context and pair sequences of length at most $T$,
 \[
 \log\frac{\det V_t^0}{\det(\alpha_{\mathrm{NTK}}I)}
 \le\overline\Gamma_T^0.
 \]
\end{enumerate}

Define
\begin{equation}
\begin{aligned}
 \mathcal E_{\mathrm{cpl}}
 &:=\mathcal E_{\mathrm{env}}\cap\mathcal E_{\mathrm{lin}}
 \cap\mathcal E_{\mathrm{id}}\cap\mathcal E_{\mathrm{opt}},\\
 \delta_{\mathrm{cpl}}
 &:=\delta_{\mathrm{env}}+\delta_{\mathrm{lin}}
 +\delta_{\mathrm{id}}+\delta_{\mathrm{opt}},\\
 \epsilon_{\mathrm{fw}}(m,T)
 &:=\epsilon_{\mathrm{lin}}(m,T)+\epsilon_{\mathrm{id}}(m,T),\\
 \epsilon_{\mathrm{cpl}}(m,T,J)
 &:=\epsilon_{\mathrm{fw}}(m,T)+\epsilon_{\mathrm{opt}}(m,T,J).
\end{aligned}
\label{eq:ntk-coupling-components}
\end{equation}
Then $\PP(\mathcal E_{\mathrm{cpl}})\ge1-\delta_{\mathrm{cpl}}$ by the union
bound. Here $\epsilon_{\mathrm{lin}}$ is the nonlinear Taylor remainder,
$\epsilon_{\mathrm{id}}$ is the argmin-stability or ridge-identification error
relative to the exact optimizer of Eq.~\eqref{eq:ntk-linearized-loss}, and
$\epsilon_{\mathrm{opt}}$ is the residual from stopping after $J$ steps.
\end{assumption}

The representer $w^\star$ and $\epsilon_{\mathrm{init}}(m)$ are selected before
online interaction and do not depend on online observations. The coupling event
contains no moving-gradient ridge equivalence or dynamic Gram comparison. Its
component bounds and the triangle inequality imply the direct nonlinear-to-ridge bound
\[
\begin{aligned}
 &\left|h(a,x;\theta_{t-1})-h(a',x;\theta_{t-1})
 -\{g_0(a,x)-g_0(a',x)\}^\top\widehat w_{t-1}^0\right|\\
 &\qquad\le\epsilon_{\mathrm{cpl}}(m,T,J).
\end{aligned}
\]
This is the finite-width coupling target required to transfer the fixed-NTK
confidence bound to the nonlinear predictor.

For almost every initialization, approximation holds for $P_X$-almost every
context. Independence of the initialization and the i.i.d. context sequence then
implies that it holds at all finitely many realized contexts almost surely. Under
exact initialization realizability, $\epsilon_{\mathrm{init}}(m)=0$.

For ReLU networks, the nonsmooth activation requires the sign-stability analysis
used by NeuralUCB and Neural Thompson Sampling. For a smooth activation, the same
role is played by uniform Hessian control in the neighborhood visited by gradient
descent. Pairwise neural bandit analysis provides the closest template because its
loss and confidence objects are also action contrasts. In either case, the width
and optimization length must control $\epsilon_{\mathrm{fw}}$ and
$\epsilon_{\mathrm{opt}}$ uniformly over the adaptive pair sequence.

Define
$\Gamma_t^0:=\log\{\det(V_t^0)/\det(\alpha_{\mathrm{NTK}}I)\}$, including
$\Gamma_0^0=0$. For $t\ge0$, let
\begin{equation}
\begin{aligned}
 \beta_t^{\mathrm{NTK}}
 &:={R_{\mathrm{ac}}}
 \sqrt{\Gamma_t^0+2\log(1/\delta)}
 +\sqrt{\alpha_{\mathrm{NTK}}}S_{\mathrm{NTK}}
 +2\epsilon_{\mathrm{init}}(m)\sqrt t,\\
 \epsilon_{m,J}
 &:=2\epsilon_{\mathrm{init}}(m)
 +\epsilon_{\mathrm{cpl}}(m,T,J).
\end{aligned}
\label{eq:ntk-radius}
\end{equation}
In particular,
$\beta_0^{\mathrm{NTK}}
=R_{\mathrm{ac}}\sqrt{2\log(1/\delta)}
+\sqrt{\alpha_{\mathrm{NTK}}}S_{\mathrm{NTK}}$.
The $2\epsilon_{\mathrm{init}}(m)\sqrt t$ term in $\beta_t^{\mathrm{NTK}}$
accumulates approximation residuals from the $t$ historical regression targets.
The separate $2\epsilon_{\mathrm{init}}(m)$ in $\epsilon_{m,J}$ controls the two
endpoints of the current candidate pair.
For a candidate pair $a,a'$, set
\begin{equation}
 r_t(a,a'):=
 \beta_{t-1}^{\mathrm{NTK}}
 \|g_0(a)-g_0(a')\|_{(V_{t-1}^0)^{-1}}
 +\epsilon_{m,J}.
 \label{eq:ntk-pair-radius}
\end{equation}

\begin{lemma}[Action-centered NTK contrast confidence]
\label{lem:ntk-contrast-confidence}
Under Assumptions~\ref{ass:iid}, \ref{ass:bounded}, \ref{ass:subgauss},
\ref{ass:ntk-cate}, and~\ref{ass:ntk-stability}, there is an event
$\mathcal C_\delta$ with probability at least $1-\delta$ such that, on
$\mathcal E_{\mathrm{cpl}}\cap\mathcal C_\delta$, simultaneously for every
$t\le T$ and $a,a'\in\A$,
\begin{equation}
\begin{aligned}
 &\left|
 \{\tau_a(X_t)-\tau_{a'}(X_t)\}
 -\{h(a,X_t;\theta_{t-1})-h(a',X_t;\theta_{t-1})\}
 \right|\\
 &\qquad\le r_t(a,a').
\end{aligned}
\label{eq:ntk-contrast-confidence}
\end{equation}
\end{lemma}

\begin{proof}
Under the original filtration and probability measure,
Proposition~\ref{prop:ntk-centered-identity} gives the martingale difference
$\nu_s$. The predictable-covariate self-normalized inequality of
\citet{abbasi2011improved} therefore defines
an event $\mathcal C_\delta$ with $\PP(\mathcal C_\delta)\ge1-\delta$ on which,
simultaneously over $t$,
\[
 \left\|\sum_{s\le t}d_s^0\nu_s\right\|_{(V_t^0)^{-1}}
 \le R_{\mathrm{ac}}
 \sqrt{\Gamma_t^0+2\log(1/\delta)}.
\]
This concentration step is not conditioned on the possibly path-dependent event
$\mathcal E_{\mathrm{cpl}}$. On the intersection
$\mathcal C_\delta\cap\mathcal E_{\mathrm{cpl}}$, which has probability at least
$1-\delta-\delta_{\mathrm{cpl}}$, initialization approximation gives
\[
 \widetilde Y_s=(d_s^0)^\top w^\star+e_s+\nu_s,
 \qquad |e_s|\le2\epsilon_{\mathrm{init}}(m),
\]
where $d_s^0$ is $\mathcal G_s$-measurable.
The deterministic approximation residual satisfies
\[
 \left\|\sum_{s\le t}d_s^0e_s\right\|_{(V_t^0)^{-1}}
 \le\|e_{1:t}\|_2
 \le2\epsilon_{\mathrm{init}}(m)\sqrt t.
\]
The first inequality follows because the nonzero eigenvalues of
$D_t(\alpha_{\mathrm{NTK}}I+D_t^\top D_t)^{-1}D_t^\top$ lie in $[0,1)$,
where $D_t$ has rows
$d_s^0{}^\top$. Finally,
$\|\alpha_{\mathrm{NTK}}w^\star\|_{(V_t^0)^{-1}}
\le\sqrt{\alpha_{\mathrm{NTK}}}S_{\mathrm{NTK}}$.
The normal equations in Eq.~\eqref{eq:ntk-ridge-update} therefore imply
\[
 \|\widehat w_t^0-w^\star\|_{V_t^0}
 \le\beta_t^{\mathrm{NTK}}.
\]
For any candidate pair, add the two initialization-approximation residuals and
apply Cauchy-Schwarz to
$\{g_0(a)-g_0(a')\}^\top(\widehat w_{t-1}^0-w^\star)$. The direct nonlinear
coupling in Assumption~\ref{ass:ntk-stability} adds
$\epsilon_{\mathrm{cpl}}(m,T,J)$, yielding
Eq.~\eqref{eq:ntk-contrast-confidence}.
\end{proof}

\subsection{Regret theorem}

Define the centered NTK effective dimension by
\begin{equation}
 \widetilde d_{\mathrm{ac}}
 :=\max\left\{1,\
 \frac{\overline\Gamma_T^0}
 {\log(1+4L_g^2T/\alpha_{\mathrm{NTK}})}\right\}.
 \label{eq:ntk-effective-dimension}
\end{equation}
The deterministic capacity $\overline\Gamma_T^0$ takes a worst-case supremum
over initialization-feature context and pair sequences. This choice avoids an
additional probabilistic complexity event for the adaptive sequence and gives a
deterministic envelope for the queue and martingale bounds.
This is the pairwise analogue of the effective dimension in NeuralUCB and Neural
Thompson Sampling. It measures the capacity of tangent contrasts over all
candidate pairs; the deterministic supremum is not restricted to the realized
active sets. It can be smaller than the effective dimension of the full
un-centered action-context kernel. The bound depends on this information gain
rather than directly on the raw parameter dimension. Under the direct coupling
event, width enters only through the initialization kernel and the finite-width
error.

\begin{theorem}[Conditional regret for the online neural-tangent variant]
\label{thm:ntk-regret}
Suppose Assumptions~\ref{ass:iid}, \ref{ass:bounded}, \ref{ass:subgauss},
\ref{ass:ntk-cate}, and~\ref{ass:ntk-stability} hold.
Run Algorithm~\ref{alg:ntk-cbwk} with
$\alpha_{\mathrm{NTK}}\ge4L_g^2$ and $\lambda_1\le R_h/c_{\min}$. Write
$\eta:=\eta_0/\sqrt T$ for the constant dual step used by the algorithm. Let
\begin{equation}
\begin{aligned}
 \overline\beta_T
 &:=R_{\mathrm{ac}}
 \sqrt{\overline\Gamma_T^0+2\log(1/\delta)}
 +\sqrt{\alpha_{\mathrm{NTK}}}S_{\mathrm{NTK}}
 +2\epsilon_{\mathrm{init}}(m)\sqrt T,\\
 \overline r_T
 &:=\frac{L_g\overline\beta_T}{\sqrt{\alpha_{\mathrm{NTK}}}}
 +\epsilon_{m,J},\\
 \overline\lambda_T
 &:=\frac{R_h+2\overline r_T}{c_{\min}},\\
 \Lambda_T
 &:=\max\{\lambda_1,\overline\lambda_T+\eta(c_{\max}+\rho)\},\\
 N_T^{\mathrm{aff}}
 &:=\left\lceil
 \frac{\Lambda_T/\eta+c_{\max}}{\rho}
 \right\rceil.
\end{aligned}
\label{eq:ntk-queue-range}
\end{equation}
Let $\mathcal C_\delta$ be the concentration event from
Lemma~\ref{lem:ntk-contrast-confidence}, and let $\mathcal Z_\delta$ be the
stopped Azuma-Hoeffding event defined in the proof below. With probability at least
$1-2\delta-\delta_{\mathrm{cpl}}$, the event
$\mathcal E_{\mathrm{cpl}}\cap\mathcal C_\delta\cap\mathcal Z_\delta$ holds and
the regret against the fluid LP benchmark satisfies
\begin{equation}
\begin{aligned}
 R(T)\le{}&
 3\beta_T^{\mathrm{NTK}}\sqrt{2T\Gamma_T^0}
 +3\epsilon_{m,J}T\\
 &+\frac{\lambda_1^2}{2\eta}
 +\frac{\eta T(c_{\max}+\rho)^2}{2}\\
 &+C\{R_{\mathrm{NTK}}+\Lambda_T(c_{\max}+\rho)\}\\
 &\qquad\times\sqrt{T\log(1/\delta)}
 +R_{\mathrm{NTK}}N_T^{\mathrm{aff}},
\end{aligned}
\label{eq:ntk-regret-exact}
\end{equation}
for a numerical constant $C$. The affordable-action rule enforces
$\sum_{t=1}^T c(X_t,A_t)\le B$ on every sample path.
\end{theorem}

\begin{proof}
Let
$s_t^\star(a)=\tau_a(X_t)-\lambda_t c(X_t,a)$, and let $a_t^\star$ maximize this
score over $\A_t^{\mathrm{aff}}$. On the confidence event of
Lemma~\ref{lem:ntk-contrast-confidence}, no action can eliminate
$a_t^\star$, so $a_t^\star\in\mathcal B_t^{\mathrm{NTK}}$. The sampled action is
one endpoint of the farthest pair and also lies in the active set. Its active-set
inequality gives
\[
 \widehat s_t(a_t^\star)-\widehat s_t(A_t)
 \le2r_t(a_t^\star,A_t).
\]
Because costs are known exactly, the pairwise confidence lemma directly gives
\[
 \left|\{s_t^\star(a_t^\star)-s_t^\star(A_t)\}
 -\{\widehat s_t(a_t^\star)-\widehat s_t(A_t)\}\right|
 \le r_t(a_t^\star,A_t).
\]
These are a single pairwise error and the active-set error, rather than two
separate endpoint errors. Hence
\[
 s_t^\star(a_t^\star)-s_t^\star(A_t)
 \le3r_t(a_t^\star,A_t).
\]
By the farthest-pair definition,
\[
 \|g_0(a_t^\star)-g_0(A_t)\|_{(V_{t-1}^0)^{-1}}
 \le\|d_t^0\|_{(V_{t-1}^0)^{-1}}.
\]
For a singleton active set, both sides are zero; otherwise the inequality follows
from maximality over distinct pairs.
Since $\|d_t^0\|_2\le2L_g$ and
$V_{t-1}^0\succeq\alpha_{\mathrm{NTK}}I$,
$\|d_t^0\|_{(V_{t-1}^0)^{-1}}^2\le1$. The elliptical-potential identity gives
\[
 \sum_{t=1}^T\|d_t^0\|_{(V_{t-1}^0)^{-1}}^2
 \le2\log\frac{\det V_T^0}{\det(\alpha_{\mathrm{NTK}}I)}
 =2\Gamma_T^0.
\]
Cauchy-Schwarz and monotonicity of the radius therefore yield
\begin{equation}
 \sum_{t=1}^T
 \{s_t^\star(a_t^\star)-s_t^\star(A_t)\}
 \le3\beta_T^{\mathrm{NTK}}\sqrt{2T\Gamma_T^0}
 +3\epsilon_{m,J}T.
 \label{eq:ntk-primal-bound}
\end{equation}
This is the neural online term. Through Eq.~\eqref{eq:ntk-radius}, it includes the
statistical information gain, the NTK RKHS norm, and the finite-width errors.
Assumption~\ref{ass:ntk-stability} gives
$\Gamma_t^0\le\overline\Gamma_T^0$ and hence
$\beta_t^{\mathrm{NTK}}\le\overline\beta_T$ for every $t\le T$.

It remains to control the budget terms. If $\lambda_t>\overline\lambda_T$, then
for every priced action $a$,
\[
 \widehat s_t(a_0)-\widehat s_t(a)
 =\lambda_t c(X_t,a)-h(a,X_t;\theta_{t-1})
 >2r_t(a_0,a).
\]
Thus all priced actions are eliminated, the algorithm plays $a_0$, and the queue
decreases. Any other round raises the queue by at most
$\eta(c_{\max}+\rho)$. Induction proves $0\le\lambda_t\le\Lambda_T$.
The standard squared-queue calculation gives
\[
 -\sum_{t=1}^T\lambda_t\{c(X_t,A_t)-\rho\}
 \le\frac{\lambda_1^2}{2\eta}
 +\frac{\eta T(c_{\max}+\rho)^2}{2}.
\]
The nonnegative queue update also gives the deterministic prefix bound
\[
 \sum_{s\le t}c(X_s,A_s)
 \le t\rho+\frac{\Lambda_T}{\eta}.
\]
Before round $t$, this bound implies
\[
 B-S_{t-1}
 \ge (T-t+1)\rho-\frac{\Lambda_T}{\eta}.
\]
Whenever the right-hand side is at least $c_{\max}$, every action is affordable.
By the definition of $N_T^{\mathrm{aff}}$, restricted affordability can therefore
occur only during the final $N_T^{\mathrm{aff}}$ rounds. The free control is always affordable and has
adjusted score zero. Replacing the global Lagrangian maximizer by the affordable
maximizer thus costs at most $R_{\mathrm{NTK}}$ on each restricted round and at
most $R_{\mathrm{NTK}}N_T^{\mathrm{aff}}$ in total.

Finally, define
\[
\begin{aligned}
 \Phi_t(a,\lambda)
 &:=\tau_a(X_t)-\lambda\{c(X_t,a)-\rho\},\\
 \mathcal L(\lambda)
 &:=\E_X\!\left[\max_{a\in\A}
 \{\tau_a(X)-\lambda(c(X,a)-\rho)\}\right].
\end{aligned}
\]
Weak duality gives $\OPT_{\mathrm{LP}}(\rho)\le\mathcal L(\lambda_t)$. Since
$X_t$ is independent of $\mathcal H_{t-1}^{\mathrm N}$ and $\lambda_t$ is
$\mathcal H_{t-1}^{\mathrm N}$-measurable,
\[
 Z_t:=\mathcal L(\lambda_t)
 -\max_{a\in\A}\Phi_t(a,\lambda_t)
 \]
is a martingale difference. Define the stopping time
$\zeta=\inf\{t:\lambda_t>\Lambda_T\}$, with $\inf\varnothing=T+1$, and the
stopped martingale
\[
 \mathcal M_T^{\zeta}:=\sum_{t=1}^T
 \mathbf 1\{t<\zeta\}
 \left\{\mathcal L(\lambda_t)
 -\max_{a\in\A}\Phi_t(a,\lambda_t)\right\}
\]
whose increments are bounded in absolute value by the deterministic quantity
$2\{R_{\mathrm{NTK}}+\Lambda_T(c_{\max}+\rho)\}$. Because $\lambda_t$ is
$\mathcal H_{t-1}^{\mathrm N}$-measurable and $\Lambda_T$ is deterministic, the stopped
process is adapted to the original filtration; its concentration is not
conditioned on $\mathcal E_{\mathrm{cpl}}$. Let
$\mathcal Z_\delta$ be its Azuma-Hoeffding event. Then
$\PP(\mathcal Z_\delta)\ge1-\delta$. On $\mathcal E_{\mathrm{cpl}}$, the queue
bound above implies $\zeta=T+1$, so
$\mathcal M_T^{\zeta}=\sum_{t=1}^T Z_t$. The identity
$\tau_{A_t}(X_t)=\Phi_t(A_t,\lambda_t)
+\lambda_t\{c(X_t,A_t)-\rho\}$ therefore gives
\begin{equation}
\begin{aligned}
 R(T)\le{}&
 \sum_{t=1}^T\left\{
 \max_{a\in\A}\Phi_t(a,\lambda_t)
 -\Phi_t(A_t,\lambda_t)\right\}\\
 &-\sum_{t=1}^T\lambda_t\{c(X_t,A_t)-\rho\}
 +\mathcal M_T^{\zeta}.
\end{aligned}
\label{eq:ntk-master}
\end{equation}
On $\mathcal Z_\delta$, the last term is bounded by the fourth term of
Eq.~\eqref{eq:ntk-regret-exact}. Combining this event with
Lemma~\ref{lem:ntk-contrast-confidence}, Eq.~\eqref{eq:ntk-primal-bound}, the
deterministic dual bound, and the affordable-set comparison proves the result.
The definition of the affordable set gives pathwise feasibility.
\end{proof}

\begin{corollary}[Effective-dimension form]
\label{cor:ntk-effective-regret}
Under the hypotheses of Theorem~\ref{thm:ntk-regret}, suppose $L_g$, $R_h$,
$R_{\mathrm{NTK}}$, $\alpha_{\mathrm{NTK}}$, $S_{\mathrm{NTK}}$, $\Xi$,
$\eta_0$, $\rho$, $c_{\min}^{-1}$, and $c_{\max}$ are controlled independently
of $T$. On
$\mathcal E_{\mathrm{cpl}}\cap\mathcal C_\delta\cap\mathcal Z_\delta$, which
has probability at least $1-2\delta-\delta_{\mathrm{cpl}}$,
Eq.~\eqref{eq:ntk-regret-exact} implies
\begin{equation}
\begin{aligned}
 R(T)=\widetilde O\Big(&
 \widetilde d_{\mathrm{ac}}\sqrt T
 +\epsilon_{\mathrm{init}}(m)T
 \sqrt{\widetilde d_{\mathrm{ac}}\log T}\\
 &+\{\epsilon_{\mathrm{fw}}(m,T)
 +\epsilon_{\mathrm{opt}}(m,T,J)\}T
 \Big).
\end{aligned}
\label{eq:ntk-regret-effective}
\end{equation}
A sufficient set of finite-width and optimization rates for preserving the
statistical order is
\[
 \epsilon_{\mathrm{init}}(m)
 =\widetilde O\!\left(\sqrt{\frac{\widetilde d_{\mathrm{ac}}}{T}}\right),
 \qquad
 \epsilon_{\mathrm{fw}}(m,T)+\epsilon_{\mathrm{opt}}(m,T,J)
 =\widetilde O\!\left(\frac{\widetilde d_{\mathrm{ac}}}{\sqrt T}\right)
 .
\]
Under these sufficient rates, the regret preserves the
$\widetilde O(\widetilde d_{\mathrm{ac}}\sqrt T)$ statistical order. The weaker
conditions
$\epsilon_{\mathrm{init}}(m)=o((\widetilde d_{\mathrm{ac}}\log T)^{-1/2})$
and $\epsilon_{\mathrm{fw}}(m,T)+\epsilon_{\mathrm{opt}}(m,T,J)=o(1)$,
together with $\widetilde d_{\mathrm{ac}}\log T=o(\sqrt T)$, suffice for
sublinear regret at a fixed confidence level. The dimension condition ensures
that the statistical contribution is itself $o(T)$.
\end{corollary}

\begin{proof}
Assumption~\ref{ass:ntk-stability} and
Eq.~\eqref{eq:ntk-effective-dimension} imply
$\Gamma_T^0=O(\widetilde d_{\mathrm{ac}}\log T)$. Substitute this relation into
Eqs.~\eqref{eq:ntk-radius}, \eqref{eq:ntk-queue-range}, and
\eqref{eq:ntk-regret-exact}. The statistical product
$\sqrt{\Gamma_T^0}\sqrt{T\Gamma_T^0}$ is
$O(\Gamma_T^0\sqrt T)$. Initialization approximation contributes
$O(\epsilon_{\mathrm{init}}T\sqrt{\Gamma_T^0})$, and direct nonlinear coupling
contributes $O(\epsilon_{\mathrm{cpl}}T)$. Because
$\widetilde d_{\mathrm{ac}}\ge1$, the queue range is at most
$\widetilde O(\sqrt{\widetilde d_{\mathrm{ac}}}
+\epsilon_{\mathrm{init}}\sqrt T+\epsilon_{\mathrm{cpl}})$.
Multiplying this quantity by $\sqrt T$ in the master-martingale and
affordable-set terms is dominated by the three displayed orders in
Eq.~\eqref{eq:ntk-regret-effective}. This proves that equation; the two rate
statements follow by direct comparison with
$\widetilde d_{\mathrm{ac}}\sqrt T$ and $T$.
\end{proof}

\end{document}